\documentclass[journal]{IEEEtran}
\usepackage[english]{babel}
\usepackage{hyperref}
\usepackage{balance}
\usepackage{graphicx} 
\graphicspath{{ {./images/} }}
\usepackage{epsfig} 
\usepackage{amsmath} 
\usepackage{amssymb}  
\usepackage{bm}
\usepackage{color}
\usepackage{caption}
\usepackage{cite}
\usepackage{rotating}
\usepackage{multirow}
\usepackage{algorithm}
\usepackage{algpseudocode}
\usepackage[switch]{lineno}
\usepackage{textcomp}
\def\BibTeX{{\rm B\kern-.05em{\sc i\kern-.025em b}\kern-.08em
    T\kern-.1667em\lower.7ex\hbox{E}\kern-.125emX}}
\title{\LARGE \bf Adaptive Koopman Embedding for Robust Control of Complex Nonlinear Dynamical Systems}

\author{~Rajpal~Singh$^{*\dag}$, Chandan~Kumar~Sah$^\dag$~and~Jishnu~Keshavan
\thanks{\dag These authors have contributed equally to the present work.}
\thanks{ This research was supported in part by the SERB Core Research Grant.}
\thanks{ The authors are with the Department of Mechanical Engineering, Indian Institute of Science, Bangalore, Karnataka~560012, India (email:  rajpalsingh@iisc.ac.in, chandanks@iisc.ac.in, kjishnu@iisc.ac.in).}
}

\begin{document}

\maketitle
\begin{abstract}
The discovery of linear embedding is the key to the synthesis of linear control techniques for nonlinear systems. In recent years, while Koopman operator theory has become a prominent approach for learning these linear embeddings through data-driven methods, these algorithms often exhibit limitations in generalizability beyond the distribution captured by training data and are not robust to changes in the nominal system dynamics induced by intrinsic or environmental factors. To overcome these limitations, this study presents an adaptive Koopman architecture capable of responding to the changes in system dynamics online. The proposed framework initially employs an autoencoder-based neural network that utilizes input-output information from the nominal system to learn the corresponding Koopman embedding offline. Subsequently, we augment this nominal Koopman architecture with a feed-forward neural network that learns to modify the nominal dynamics in response to any deviation between the predicted and observed lifted states, leading to improved generalization and robustness to a wide range of uncertainties and disturbances compared to contemporary methods. Extensive tracking control simulations, which are undertaken by integrating the proposed scheme within a Model Predictive Control framework, are used to highlight its robustness against measurement noise, disturbances, and parametric variations in system dynamics.


\end{abstract}

\section{Introduction}
\label{sec: intro}
The control of complex nonlinear dynamical systems remains a critical problem across diverse scientific and engineering disciplines~\cite{strogatz2018nonlinear}. Unfortunately, a unifying mathematical framework for the control of such complex systems remains elusive. The well-developed field of linear control theory offers a robust toolbox for analyzing and controlling linear systems. Consequently, there is a significant appeal in developing methods that represent nonlinear dynamical systems within a linear framework. Koopman operator theory~\cite{koopman1931hamiltonian} has emerged as one of the prominent methods for obtaining linear representations of any nonlinear systems. Unlike local methods like Taylor series linearization, Koopman operator theory linearly encodes the system dynamics throughout the entire workspace and, hence, offers a significantly superior prediction performance~\cite{korda2018linear}. 

In particular, Koopman operator theory leverages nonlinear observable functions of the system states that reside within an infinite-dimensional Hilbert space to represent system dynamics. By lifting the original dynamics through these infinite-dimensional observables, the ensuing system evolves linearly via infinite-dimensional Koopman operator~\cite{mauroy2020koopman,brunton2021modern}. However, the practical application of Koopman operator theory is often impeded by its infinite dimensionality and the absence of a comprehensive mathematical framework for computing the operator accurately. Nevertheless, data-driven methods such as Dynamic Mode Decomposition (DMD)~\cite{rowley2009spectral}, Extended DMD (EDMD)~\cite{williams2015data}, and Sparse Identification of Nonlinear Dynamics (SINDY)~\cite{brunton2016discovering} have been proposed to compute a finite-dimensional approximation of the Koopman operator for autonomous systems, while DMD with Control (DMDc)~\cite{proctor2016dynamic}, EDMD with control (EDMDc), and SINDY with control (SINDYc)~\cite{brunton2016sparse} are their extensions to control-affine systems. These linear representations frequently suffer from reduced prediction accuracy, particularly when dealing with more complex systems. Hence, bilinear Koopman representations have been suggested~\cite{goswami2017global,bruder2021advantages} to better approximate nonlinear systems in contrast with a purely linear one. Although this approach enhances prediction, it remains incompatible with linear control methods. Moreover, the selection of basis/observable functions significantly impacts the learning performance of both bilinear and linear Koopman representations. While traditionally performed through manual selection, this approach becomes increasingly laborious for complex systems and often yields suboptimal results. Consequently, learning-based algorithms such as dictionary learning~\cite{folkestad2021diction}, autoencoders~\cite{lusch2018deep}, variational autoencoders~\cite{murata2019finding}, and invertible neural networks~\cite{meng2024koopman} have been proposed to automate the identification of an optimal basis set. The efficacy of these learning frameworks in generating high-fidelity linear/bilinear Koopman models has fostered their application in the modeling and control of robotic systems such as manipulators~\cite{ sah2024real,bruder2021advantages}, quadrotors~\cite{folkestad2021diction,folkestadEpisodic, folkestad2022koopnet}, and soft robots~\cite{bruder2019nonlinear, haggerty2020modeling, castano2020control}.


The performance of model-based control methods is heavily influenced by the accuracy of the underlying model, underscoring the significance of high-fidelity system identification algorithms.
Since Koopman-based algorithms rely on data for learning, they suffer from the limitations of data-driven algorithms. For instance, the validity of the learned model is confined to the specific region of the configuration space captured in the training data. Hence, the learned model doesn't generalize well outside the space captured by training data. However, many real-world scenarios involve systems with dynamics that can vary over time. Thus, any changes in the system's dynamics resulting from variations in system parameters or environmental conditions render the learned model obsolete. This requires fresh data collection from the modified system and subsequent retraining to account for any unmodeled dynamics and/or disturbances, which is computationally expensive and time-consuming. This significantly restricts the applicability of most previous Koopman frameworks. Hence, some recent studies have been proposed to improve the accuracy of the previously learned system models; for instance, Jacobian-Regularized Dynamic-Mode Decomposition (JDMD) proposed in~\cite{JDMD2023} offers improved efficiency over traditional Koopman approaches based on DMD by leveraging Jacobian information from a faulty prior theoretical model of the system. Similarly, Koopman models are retrained offline for a better control performance in~\cite{uchida2023control}. However, these Koopman architectures do not extend to include any adaptive module online, which allows learning unmodeled dynamics or changes in the system dynamics during deployment without the need to retrain it from scratch. 

To mitigate this problem, we propose an adaptive Koopman architecture that leverages both offline and online learning to account for variations in system dynamics. The nominal Koopman model consists of an autoencoder-based neural structure that uses input-output information from the nominal system to learn the corresponding linear/bilinear Koopman model offline. This offline structure is then augmented with a feed-forward neural network that learns any model uncertainties or disturbances affecting system dynamics online.
Owing to this online learning structure, the proposed architecture demonstrates remarkable versatility in adapting to complex variations in system dynamics, thus rendering it sample-efficient with respect to how such large variations in dynamics are implicitly captured through training data.

 In addition, it is resource-intensive and time-consuming to generate large datasets for complex systems to train the neural networks effectively. The proposed adaptive algorithm mitigates this problem by continuously gathering data online, thereby enabling it to correct faulty/poorly trained Koopman models and enhance the accuracy of the learned model, reducing the burden of requiring a vast offline dataset and facilitating more efficient model learning for complex systems. Consequently, online adaptation allows even poorly trained inaccurate Koopman models to acquire high prediction accuracy.
 Further, the proposed adaptive architecture empowers the linear Koopman model to achieve a prediction performance similar to or even better than their nominal bilinear counterparts, enabling full integration of these models within linear control frameworks. In this study, the proposed adaptive Koopman model is paired with a Model Predictive Control (MPC) scheme for the control of nonlinear systems. This online adaptation method also makes the model more amenable to control, as suggested in~\cite{uchida2023control}.

Our work stands alone in the Koopman literature as it leverages an adaptive network to learn changing dynamics/uncertainties online, resulting in more accurate models suitable for precise control. While our research is closely related to the JDMD algorithm proposed in~\cite{JDMD2023}, it is worth noting that JDMD relies on a pre-existing mathematical model and operates offline. Consequently, it lacks robustness against unmodeled dynamics and disturbances. Unlike previous works, our adaptive algorithm is designed to operate beyond the workspace captured by the training data, which is a significant advantage over conventional data-driven approaches. We demonstrate tracking control with a wide range of robotic systems to highlight the superior performance and robustness of the proposed scheme.




\section{Results}
\subsection{Koopman Preliminaries}\label{prelim}
\begin{figure*}[ht!]
     \centering
     \includegraphics[width=\textwidth]{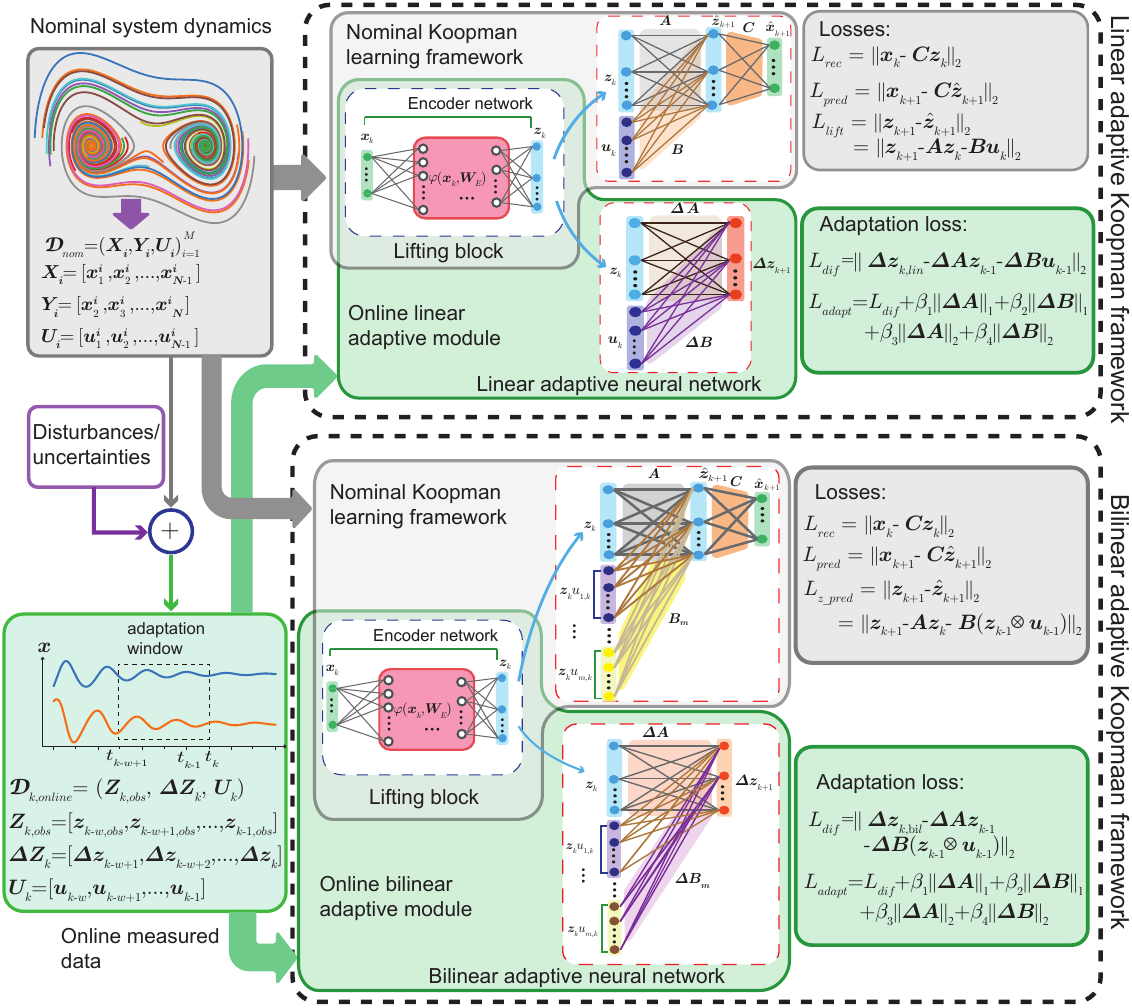}
         \caption{\textbf{Illustration of the proposed adaptive embedding along with their neural network architectures}. The proposed architecture combines online and offline learning. An offline model (shown in grey-colored blocks) is initially leveraged to train a nominal model by minimizing $L_{nom}$. The dataset $\mathcal{D}_{nom}$, generated from nominal system dynamics, consists of snapshot matrices of states $\boldsymbol{X}_i$ and $\boldsymbol{Y}_i$ together with control inputs $\boldsymbol{U}_i$ for each trajectory $i$, used to train the nominal model. The online network (shown in green colored blocks) consists of the pre-trained lifting block and an adaptive neural network block. The adaptive network is employed to update the nominal Koopman model by adapting to any change in dynamics that happens post-training or unmodeled dynamics unaccounted for in the training dataset. The online adaptation module minimizes the adaptation losses over the window of $w$ latest datasets collected online during deployment. For time-step $k$, the online dataset $\mathcal{D}_{k, online}$ includes snapshot matrices $\boldsymbol{Z}_{k,obs},\boldsymbol{\Delta Z}_{k,obs},$ and $\boldsymbol{U}_{k,obs}$ which contain past $w$ values of the actual lifted state (obtained after lifting the measured state $\boldsymbol{x}_{obs}$) of the system~($\boldsymbol{z}_{k,obs}$),  prediction error~($\boldsymbol{\Delta{z}}_k = \boldsymbol{{z}}_{k,obs} - \boldsymbol{\hat{z}}_k$), and control inputs~($\boldsymbol{u}_k$).}
    \label{fig:adap_koop}
\end{figure*}
To give context to our work, we first summarize the fundamentals of the Koopman Operator theory. Consider a nonlinear autonomous system described as:
\begin{eqnarray}
    \label{eq:base_non_linear_sytem}
    \boldsymbol{\dot{x}} = \boldsymbol{f}(\boldsymbol{x}),
\end{eqnarray}
where the state of the system, $\boldsymbol{x} \in \mathbb{X} \subset \mathbb{R}^{n}$ and $\boldsymbol{f}:\mathbb{X} \to \mathbb{X}$ are assumed to be Lipschitz continuous on $\mathbb{X}$. The corresponding discrete dynamics can be given as, $\boldsymbol{{x}}_{k+1} = \boldsymbol{S}(\boldsymbol{x}_k)$. where $\boldsymbol{S}:\mathbb{X} \to \mathbb{X}$ represents the flow map and $\boldsymbol{x}_k$ describes the state at $k^{th}$ time step. The seminal study in~\cite{koopman1931hamiltonian} offers a distinct perspective for analyzing Hamiltonian systems by their transformations within a Hilbert space. This approach departs from the traditional focus on state evolution and instead emphasizes the dynamics of observable functions, $\sigma: \mathbb{X} \to \mathbb{C}$, within a Hilbert space, whose evolution is described by the continuous Koopman operator, $\mathcal{K}: \mathbb{C} \to \mathbb{C}$ as
\begin{eqnarray}    
    \mathcal{K}\circ \sigma{(\boldsymbol{x}_k)} =  \sigma\circ \boldsymbol{S}(\boldsymbol{x}_k) = \sigma(\boldsymbol{x}_{k+1}),  \nonumber
\end{eqnarray}
where $\circ$ represents functional composition. Note that the evolution of the observables in Koopman space is linear. However, the Koopman operator must be infinite-dimensional for an exact Koopman representation~\cite{mauroy2020koopman}.

The same theoretical concepts can also be extended to a system with control inputs. Consider a control-affine system of the form:
\begin{eqnarray}
        \label{eq:base_affine}
    \boldsymbol{\dot{x}} = \boldsymbol{f}_0(\boldsymbol{x}) + \sum_{i=1}^{m}\boldsymbol{f}_i(\boldsymbol{x}){u}_{i},
\end{eqnarray}
where $\boldsymbol{u} \in \mathbb{R}^m$ represents the control input to the system, and $\boldsymbol{f}_0$ and $\boldsymbol{f}_i$ represent the drift and state-dependent control vectors respectively.
The importance of control-affine systems stems from their wide applicability across diverse engineering and science fields such as robotics~\cite{folkestad2022koopnet,sah2024real,morgansen2001nonlinear}, population ecology~\cite{tian2011global}, cellular signaling \cite{sun2016mitogen}, chemical reactor optimization ~\cite{maya2003controllability}, and quantum mechanics~\cite{ boscain2002k+,khaneja2002sub}.

The system~(\ref{eq:base_affine}) admits an exact finite dimensional bilinear representation in a Hilbert space when Koopman eigenfunctions (KEFs), $\phi(\cdot)$, are employed as observable functions as given below: 
\begin{eqnarray}
\label{eq:bilinear_continious}
    \boldsymbol{\dot{z}} = \boldsymbol{A_c} \boldsymbol{z} + \boldsymbol{B}_{\boldsymbol{c}} (\boldsymbol{z} \otimes \boldsymbol{u}), \boldsymbol{x} = \boldsymbol{C}\boldsymbol{z},
\end{eqnarray}
where $\boldsymbol{z} = \boldsymbol{T}(\boldsymbol{x}) = 
[{\phi}_1(\boldsymbol{x}),{\phi}_2(\boldsymbol{x}),...,{\phi}_{p}(\boldsymbol{x})]^{T}$, $\boldsymbol{A_c} \in \mathbb{R}^{p \times p}$, $\boldsymbol{B_c} \in \mathbb{R}^{p \times mp}$ and $\boldsymbol{C} \in \mathbb{R}^{n \times p}$ are Koopman model matrices, $p$ is the dimension of the Koopman invariant subspace, and $\otimes$ represents the Kronecker product. A linear Koopman representation can similarly be given as:
\begin{eqnarray}
\label{eq:linear_continious}
\dot{\boldsymbol{z}} = \boldsymbol{A}_{c}\boldsymbol{{z}}{+} \boldsymbol{B}_{c}\boldsymbol{u},  \boldsymbol{x} = \boldsymbol{C}\boldsymbol{z},
\end{eqnarray}
where $\boldsymbol{B_c} \in \mathbb{R}^{p \times m}$ and $\boldsymbol{C} \in \mathbb{R}^{n \times p}$. A rigorous theoretical analysis of the Koopman representation of control-affine systems using the KEF basis is given in section~\ref{sec:Koop_cont_affine}.

To learn Koopman embeddings, we leverage the well-established function approximation capabilities of neural networks. This inherently automates the selection of an appropriate set of basis functions within the Koopman framework, facilitating its realization through an autoencoder architecture as shown in Fig.~\ref{fig:adap_koop}, which is referred to as a nominal linear/bilinear Koopman neural network. The autoencoder is employed to learn a set of suitable lifting functions~$\boldsymbol{\varphi}(\cdot)$ such that $\boldsymbol{{z}}_{k} = \boldsymbol{\varphi}(\boldsymbol{x}_k)$. Note that the lifting function~$\boldsymbol{\varphi}(\cdot)$ discovered offline by the autoencoder is an approximation of the KEF-based observable function $\boldsymbol{T}(\cdot)$. This autoencoder is followed by a set of dense layers that are employed to discover the Koopman model matrices  $\boldsymbol{A}, \, \boldsymbol{B}$, and $\boldsymbol{C}$.  Further details of the architecture, the definition of losses, and the training process can be found in Section~\ref{sec:neural_arc}. The linear/bilinear Koopman models learned for the nominal dynamics~(\ref{eq:base_affine}) can be expressed  as:

\begin{align}
    &\label{eq:nom_lin}
    \boldsymbol{\hat{z}}_{k{+}1} {=} \boldsymbol{A}\boldsymbol{z}_k {+} \boldsymbol{B}\boldsymbol{u}_k, \,\, \boldsymbol{\hat{x}}_{k+1} {=} \boldsymbol{C}\boldsymbol{\hat{z}}_{k+1},\\
    &\label{eq:nom_bil}
    \boldsymbol{\hat{z}}_{k{+}1} {=} \boldsymbol{A}\boldsymbol{z}_k {+} \boldsymbol{B}(\boldsymbol{{z}}_k\otimes\boldsymbol{u}_k), \,\, \boldsymbol{\hat{x}}_{k+1} {=} \boldsymbol{C}\boldsymbol{\hat{z}}_{k+1},
\end{align}
where $\boldsymbol{\hat{z}}_{k{+}1}$ and $\boldsymbol{\hat{x}}_{k+1}$ are lifted and base state as predicted by the nominal Koopman model at $(k+1)^{th}$ time step. Note that Koopman models~(\ref{eq:nom_lin}) and~(\ref{eq:nom_bil}) are the discrete-time equivalents of the continuous-time Koopman models~(\ref{eq:linear_continious}) and~(\ref{eq:bilinear_continious}) respectively.
\subsection{Adaptive Koopman Module} \label{sec:adap_koop}
Now, we consider the synthesis of an adaptive Koopman model. To this end, consider a modified control-affine~(\ref{eq:base_affine}) system as
\begin{eqnarray}
\label{eq:affine_mod}
    \boldsymbol{\dot{x}} = \boldsymbol{f_0}(\boldsymbol{x}) + \boldsymbol{\tilde{f}_0}(\boldsymbol{x})+ \sum_{i=1}^{m} (\boldsymbol{f}_i(\boldsymbol{x}) + \boldsymbol{\tilde{f}}_i(\boldsymbol{x}))u_i,
\end{eqnarray}
where $\boldsymbol{\tilde{f}_0}$ and  $\boldsymbol{\tilde{f}_i}$ represent the change in system dynamics induced by intrinsic/environmental factors. Alternatively, due to poor training or a limited amount of data, system~(\ref{eq:base_affine}) can be considered as the incomplete dynamics captured by the nominal Koopman model, while system~(\ref{eq:affine_mod}) may be considered as the actual dynamics of the system. This work hinges on the assumption that any deviations in the actual system dynamics, captured by  $\boldsymbol{\tilde{f}_0}$ and  $\boldsymbol{\tilde{f}_i}$, lead to corresponding variations in the Koopman representation. Mathematically, this translates to adjustments in the model matrices $\boldsymbol{A}$ and  $\boldsymbol{B}$, represented by $\boldsymbol{\Delta A}$ and $\boldsymbol{\Delta B}$, respectively. Section~\ref{sec:adaptation_theory} provides theoretical analysis establishing the conditions for this relationship between system dynamics deviations and their Koopman representations. Consequently, under these assumptions, system~(\ref{eq:affine_mod}) can be expressed with the following linear Koopman representation:
\begin{eqnarray}
\label{eq:adap_lin}
\boldsymbol{z}_{k{+}1} {=} (\boldsymbol{A} + \Delta\boldsymbol{A})\boldsymbol{z}_{k} {+} (\boldsymbol{B} + \Delta\boldsymbol{B})\boldsymbol{u}_k, \, \,\boldsymbol{x}_k {=} \boldsymbol{C}\boldsymbol{z}_k.
\end{eqnarray}
Note here that $\boldsymbol{{z}}_{k{+}1}$ is the lifted state predicted at $(k+1)^{th}$ time-step by the adaptive Koopman model and should be ideally equal to the lifted state obtained by lifting the actual state observed for system~(\ref{eq:affine_mod}), i.e., $\boldsymbol{z}_{k{+}1} = \boldsymbol{z}_{k{+}1,obs} = \boldsymbol{\varphi}(\boldsymbol{x}_{k{+}1,obs}) $. Hence, we make adjustments to the nominal Koopman models such that the lifted states predicted by the nominal Koopman network $\boldsymbol{\hat{z}}$ align with the lifted states corresponding to the observed states $\boldsymbol{x}_{obs}$ of the actual dynamics~(\ref{eq:affine_mod}). From Eqs.~(\ref{eq:nom_lin}) and~(\ref{eq:adap_lin}), we get
\begin{eqnarray}
    \label{eq:adapt_lin_equation}
    \boldsymbol{\Delta z}_{k,lin} = \boldsymbol{\Delta A} \boldsymbol{z}_{k-1} + \boldsymbol{ \Delta B} \boldsymbol{u}_{k-1},
\end{eqnarray}
where $\boldsymbol{\Delta z}_{k,lin}=\boldsymbol{z}_{k,obs} - \boldsymbol{\hat{z}}_{k}$ represents the prediction error at $k^{th}$
timestep for the linear Koopman representation. Then, computing the matrices $\boldsymbol{\Delta A}$ and $\boldsymbol{\Delta B}$ can be framed as the following optimization problem:
\begin{eqnarray}
    \label{eq:optim_lin_problem}
    &\min_{\boldsymbol{\Delta A}, \boldsymbol{\Delta B}} \| \boldsymbol{\Delta z}_{k,lin} {-} \boldsymbol{\Delta A} \boldsymbol{z}_{k-1} {-} \boldsymbol{ \Delta B}  \boldsymbol{u}_{k-1}\|_2 .
\end{eqnarray}
Although a computationally efficient least-squares solution exists in theory, achieving sparsity in matrices $\boldsymbol{\Delta A}$ and $\boldsymbol{\Delta B}$ for practical applications would require some form of regularization of the corresponding weights. Hence, for ease of implementation, we instead use a feed-forward neural network to learn the error system~(\ref{eq:adapt_lin_equation}).

Similarly, for the bilinear representation, the adaptive Koopman representation can be given as 
\begin{eqnarray}
\label{eq:adap_bil}
\boldsymbol{z}_{k{+}1} {=} (\boldsymbol{A} {+} \Delta\boldsymbol{A})\boldsymbol{z}_k {+} (\boldsymbol{B} {+} \Delta\boldsymbol{B})(\boldsymbol{z}_k\otimes\boldsymbol{u}_k),\,\boldsymbol{x}_k {=} \boldsymbol{C}\boldsymbol{z}_k,
\end{eqnarray}
 and the corresponding prediction error can be given as
 \begin{eqnarray}
    \label{eq:adapt_equation_bil}
     \boldsymbol{\Delta z}_{k,bil} {=} \boldsymbol{\Delta A} \boldsymbol{z}_{k-1} {+} \boldsymbol{ \Delta B}(\boldsymbol{z}_{k-1} \otimes \boldsymbol{u}_{k-1}),
\end{eqnarray}
where $\boldsymbol{\Delta z}_{k,bil}$ represents the prediction error at $k^{th}$
timestep for the bilinear representation. The corresponding optimization problem can then be written as
\begin{eqnarray}
    \label{eq:optim_bilin_problem}
    \min_{\boldsymbol{\Delta A}, \boldsymbol{\Delta B}} \| \boldsymbol{\Delta z}_{k,bil} {-} \boldsymbol{\Delta A} \boldsymbol{z}_{k-1} {-} \boldsymbol{ \Delta B}(\boldsymbol{z}_{k-1}\otimes\boldsymbol{u}_{k-1})\|_2 .
\end{eqnarray}

The neural network employed to learn the $\boldsymbol{\Delta A}$ and $\boldsymbol{\Delta B}$ matrices online is referred to as an adaptive linear/bilinear neural architecture depending on the nature of the nominal Koopman model it augments. This online adaptation is further explained in Section~\ref{sec:online_adap}. A visual overview of the complete architecture is shown in Fig.~\ref{fig:adap_koop}.

\begin{figure*}[ht!]
     \centering
     \includegraphics[width=0.95\textwidth]{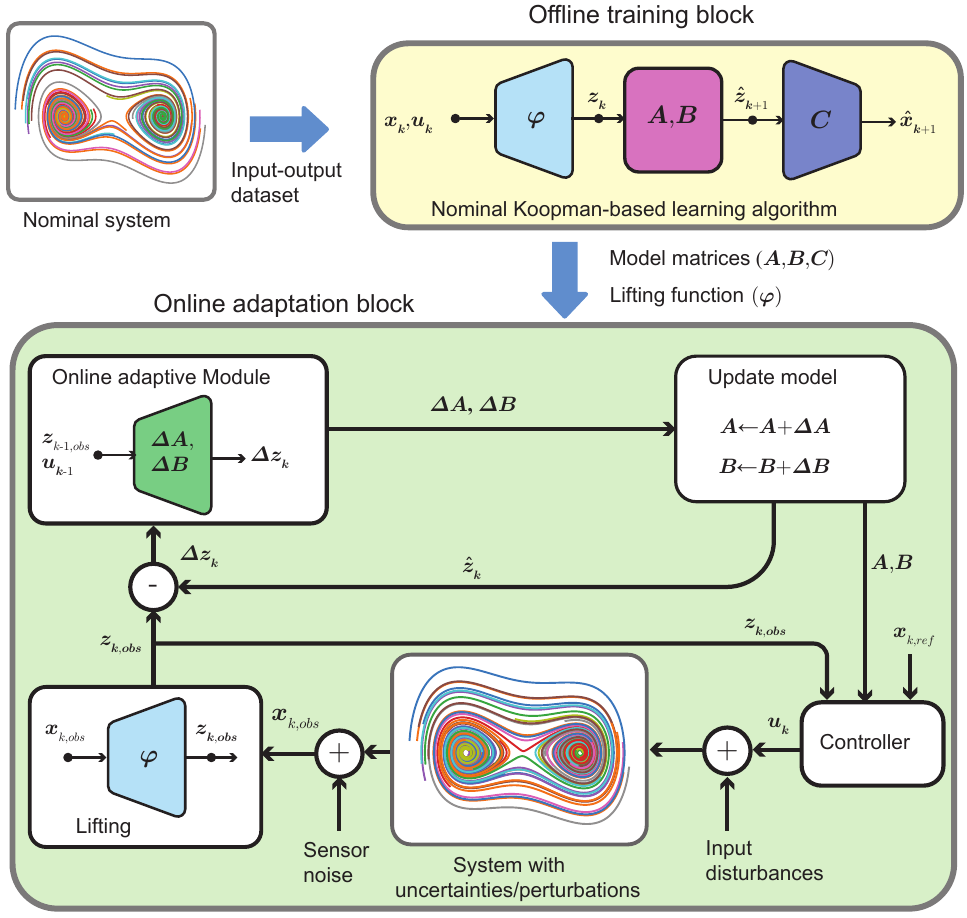}
     \caption{\textbf{Illustration of the adaptive Koopman architecture embedded in a closed loop control framework.} During the offline training phase, input-output data from the nominal system is used to train the nominal Koopman Neural networks, which, upon training, gives the model matrices $\boldsymbol{A}, \boldsymbol{B}$, $\boldsymbol{C}$ and lifting function $\boldsymbol{\varphi}(.)$. During the deployment phase, the online adaptive architecture uses the difference between predicted lifted values, $\boldsymbol{\hat{z}}_{k}$ and actual lifted values, $\boldsymbol{z}_{k, obs}$ to calculate the required correction to the Koopman matrices in the form of $\boldsymbol{\Delta A}$ and $\boldsymbol{\Delta B}$. The controller (MPC in the present study) uses the updated Koopman model matrices to compute the required control output $\boldsymbol{u}_k$ required to drive $\boldsymbol{x}_{k,obs}$ to its reference value $\boldsymbol{x}_{ref}$. This computed control input $\boldsymbol{u}_k$ is applied to the real system (with uncertainties), and the resulting state is measured.}
    \label{fig:adaptation_block}
\end{figure*}

Fig.~{\ref{fig:adaptation_block}} illustrates the overall adaptive algorithm for simultaneous adaptation and control. The input-output dataset is generated offline from the nominal system to train the Koopman neural network architecture (detailed in Section~\ref{sec:neural_arc}). This architecture yields the mapping $\boldsymbol{\varphi}(\cdot)$ and Koopman model matrices $\boldsymbol{A}$, $\boldsymbol{B}$, and $\boldsymbol{C}$.
During deployment, the sensor measurements provide the current true state, $\boldsymbol{x}_{k,obs}$, of the actual system (accounting for uncertainties/disturbances). This state is lifted to the invariant Koopman embedding space using the pre-trained lifting map, $\boldsymbol{\varphi}(\cdot)$, resulting in $\boldsymbol{z}_{k,obs}$. The online adaptive module aims to minimize the discrepancy between the measured lifted state, $\boldsymbol{z}_{k,obs}$, and its predicted counterpart, $\boldsymbol{\hat{z}}_{k}$, over a predefined adaptation window of $w$ latest data, via the minimization of the optimization problem~(\ref{eq:optim_lin_problem}) and (\ref{eq:optim_bilin_problem}) for linear and bilinear adaptive models, respectively. The solution, $\boldsymbol{\Delta A}$ and $\boldsymbol{\Delta B}$, represents the corrections made to the nominal model matrices, $\boldsymbol{A}$ and $\boldsymbol{B}$, respectively, to account for uncertainties/disturbances. Subsequently, the updated model matrices are employed in conjunction with a control strategy, such as Model Predictive Control (MPC) or Nonlinear Model Predictive Control (NMPC) (further details in Section~\ref{sec:MPC}), to compute the control input, $\boldsymbol{u}_k$ to steer the current state towards the desired reference state, $\boldsymbol{x}_{ref}$. The computed $\boldsymbol{u}_k$ is applied to the real system, and the resulting next state, $\boldsymbol{x}_{k+1,obs}$, is measured by the sensors. This iterative process continues at each time step, $k$, enabling the system to achieve the control objective while simultaneously adapting to uncertainties/disturbances. 

Although the MPC framework is employed for feedback control, it should be noted that the adaptive Koopman framework proposed in this study is controller-agnostic and can be paired with any other suitable control policy, which is a desirable attribute of the proposed scheme. 



\begin{algorithm}
\caption{Pseudo Code For Adaptive Koopman MPC}\label{alg:cap}
\textbf{Offline:}\\
\textbf{Data:} $\boldsymbol{\mathcal{D}}_{nom} = \left(\boldsymbol{X}_i, \boldsymbol{Y}_i, \boldsymbol{U}_i \right)_{i=1}^M$\;
\begin{enumerate}
    \item Calculate $\boldsymbol{\varphi}(\cdot)$, $\boldsymbol{A}$, $\boldsymbol{B}$ and $\boldsymbol{C}$ using nominal Koopman model.
\end{enumerate}
\textbf{Online:}\\
assign window: $w$\\
\textbf{For} {each discrete time step $k = 1, 2, 3, ....$ \textbf{do}}{
  \begin{itemize}
      \item $if \; k < w$:\\
      $\boldsymbol{\Delta A} = \boldsymbol{0}$,\,
        $\boldsymbol{\Delta B} = \boldsymbol{0}$
  
      \item else:
      \begin{enumerate}
        \item $\boldsymbol{z}_{k} \gets \boldsymbol{\varphi}(\boldsymbol{x}_{k})$ 
      
        \item $\boldsymbol{\mathcal{D}}_{k,online} = \{(\boldsymbol{z}_{k-w,obs}, \boldsymbol{\Delta {z}}_{k-w+1}, \boldsymbol{u}_{k-w}), \\(\boldsymbol{z}_{k-w+1,obs},\boldsymbol{\Delta {z}}_{k-w},\boldsymbol{u}_{k-w+1}),...,(\boldsymbol{z}_{k-1,obs}, \boldsymbol{\Delta{z}}_{k},\\ \boldsymbol{u}_{k-1}) \},$
        where, $\boldsymbol{\Delta z}_{i} = \boldsymbol{z}_{i,obs} - \boldsymbol{\hat{z}}_{i-1}$ for $i=k-w,...,k$ 
      \item Compute $\boldsymbol{\Delta A}$ and $\boldsymbol{\Delta B}$ using the adaptive module (\ref{eq:adapt_loss})
      \item $\boldsymbol{A} \gets \boldsymbol{A} + \boldsymbol{\Delta A}$,\,
       $\boldsymbol{B} \gets \boldsymbol{B} + \boldsymbol{\Delta B}$
      \end{enumerate}
        \item Compute the control output $\boldsymbol{u}_k$ using (\ref{eq:linear_MPC}) or (\ref{eq:bilinear_MPC}).
      \item Apply control input $\boldsymbol{u}_k$ to the system
  \end{itemize}
  \textbf{end}
}
\end{algorithm}

\subsection{Simulation Results}\label{sec:sim_results}
To assess the efficacy and generalizability of the proposed adaptive Koopman algorithm, we conduct a series of simulation studies encompassing diverse dynamical systems: a coupled pendulum, a serial manipulator, and a planar quadrotor. We assume that all these systems are fully observable. Each system initially undergoes offline training to generate a nominal Koopman model. During deployment, we introduce variations to the system parameters or inject external uncertainties/disturbances to evaluate the algorithm's capacity to adapt to these implemented variations in real-time. Throughout the simulation study, OSQP~\cite{osqp} solver is used to implement MPC and NMPC.



\subsubsection{Coupled Pendulums} \label{sec:sync_pend}
The coupled pendulum system consists of an array of simple pendulums with shared pivot points interconnected by torsion springs. This system exhibits highly nonlinear dynamics with complexity scaling exponentially with the number of pendulums. Further, it is underactuated, as the motor can directly control only one of the pendulums at the boundary, which adds to the challenge of control.
This combination of factors – high dimensionality, underactuation, and intricate, parameter-sensitive dynamics – renders traditional data-driven Koopman learning particularly arduous. Consequently, the robust adaptation capabilities embedded within our adaptive Koopman framework become crucial for successful control in this scenario. \par
In this study, we integrate the proposed adaptive Koopman-based learning with an MPC-based controller for controlled synchronization of the coupled pendulum in the presence of model uncertainties/disturbances. The problem of controlled synchronization in an underactuated coupled pendulum system represents a specific instance of reference tracking, where individual pendulums achieve a uniform state or trajectory via feedback control. This problem has been previously explored using Koopman Model Predictive Control (KMPC)
in~\cite{do2023controlled}, where Koopman-based learning extracts a linear model of the system from data, which is integrated within an MPC framework for control. The coupled pendulum system is modeled using the mechanical realization of the Frenkel-Kontorova (FK) model~\cite{braun1998nonlinear} in the form of a finite, one-dimensional array of $N$ identical pendulums as
\begin{eqnarray}
    &I \ddot{\theta}_i + m g l \sin{\theta_i} + \gamma \dot{\theta}_i - \frac{\kappa}{2}  \frac{\partial}{\partial \theta_i} \sum_{j=1}^{N-1} (\theta_{j+1} - \theta_j)^2 -\nonumber \\ 
    \label{eq:fk}
    &\frac{b}{2} \frac{\partial}{\partial \dot{\theta}_i} \sum_{j=1}^{N-1} (\dot{\theta}_{j+1} - \dot{\theta}_j)^2 = M_i,\; i=1,2,...,N,
\end{eqnarray}

where $\kappa$ is the stiffness of torsion springs coupling adjacent pendulums, $b$ and $\gamma$ are the damping coefficients, $g$ is the acceleration due to gravity, and $I$, $m$, and $l$ are the moment of inertia, mass and length of each pendulum, respectively. Since all the pendulums are identical, the suffix $i$ is omitted. $\theta_i$ and $M_i$ represent the angular displacement and  the external torques on the $i^{th}$ pendulum, respectively. Since the system is actuated by only one motor, assuming it drives the first pendulum, we have $M_1 \ne 0$ and $M_i=0$ for $i=2,\,3,...,\, N$. Eq.~(\ref{eq:fk}) can be written in state-space form using the multi-agent system formalism (refer to \cite{do2023controlled} for comprehensive details) as:


\begin{eqnarray}
    \boldsymbol{\dot{x}} = F(\boldsymbol{x}) - \frac{1}{I}(\boldsymbol{L} \otimes \boldsymbol{G} \boldsymbol{K})\boldsymbol{x} + \frac{1}{I}(\boldsymbol{d} \otimes \boldsymbol{G})u,
\end{eqnarray}
where $\boldsymbol{L} = (L_{ij})\in \mathbb{R}^{N \times N}$ is the graph Laplacian describing the topological structure of the interactions between the pendulums. The state $\boldsymbol{x} = [\boldsymbol{x}_1^{\top},...,\boldsymbol{x}_N^{\top}]^{\top}$, where $\boldsymbol{x}_i = [\theta_i, \dot{\theta}_i]^{\top}$ is the state of the $i^{th}$ pendulum. $F(\boldsymbol{x}) = [f(\boldsymbol{x}_1)^{\top},...,f(\boldsymbol{x}_N)^{\top}]^{\top}$, where $f(\boldsymbol{x}_i)$ is the uncoupled dynamics of $i^{th}$ pendulum, given by:
\begin{eqnarray}
    \label{eq:dyn_single_pend}
    f(\boldsymbol{x}_i) = 
    \begin{bmatrix}
    \boldsymbol{x}_{i,2}\\
    -({mgl}/{I}) \sin(\boldsymbol{x}_{i,1}) - ({\gamma}/{I}) \boldsymbol{x}_{i,2}
    \end{bmatrix}
\end{eqnarray}
$\boldsymbol{G} = [0,1]^{\top}$ and $\boldsymbol{K} = [\kappa, b]$ represents the torsion coupling through springs with dissipation. $\boldsymbol{d} = [1,0,...,0]^{\top} \in \mathbb{R}^{N}$, $u = M_1$. \par

For simulations, we considered a system of $N=5$ identical coupled pendulums $(\text{hence, }\boldsymbol{x} \in \mathbb{R}^{10})$ with the following dynamic parameters: $m = 0.017$ kg, $l = 0.1$ m, $\gamma = 3.1956332 \times 10^{-4}$ Nms rad$^{-1}$, $b = 6.819 \times 10^{-4}$ Nms rad$^{-1}$, and $k = 0.079$ Nm$^{-1}$. The control loop is operated at $100$ Hz, emulating realistic experimental conditions. \par

\textbf{\emph{Adaptation to Parametric Uncertainties:}}
To evaluate the performance of the proposed adaptive Koopman architecture against conventional non-adaptive approaches, we conduct a comprehensive suite of simulation studies.
Specifically, we simulate parametric uncertainties by either increasing or decreasing each constant parameter $(I,m,l,\gamma,k, \text{and } b)$ at random by $\delta\%$, where $\delta \in \{5, 10, 15, 20, 25, 30, 35, 40\}$. 
Additionally, we simulate sensor noise by corrupting the measured states with Gaussian noise at different amplitudes, characterized by various signal-to-noise (SNR) ratios. 

\begin{figure*}[ht!]
     \centering
     \includegraphics[width=0.9\textwidth]{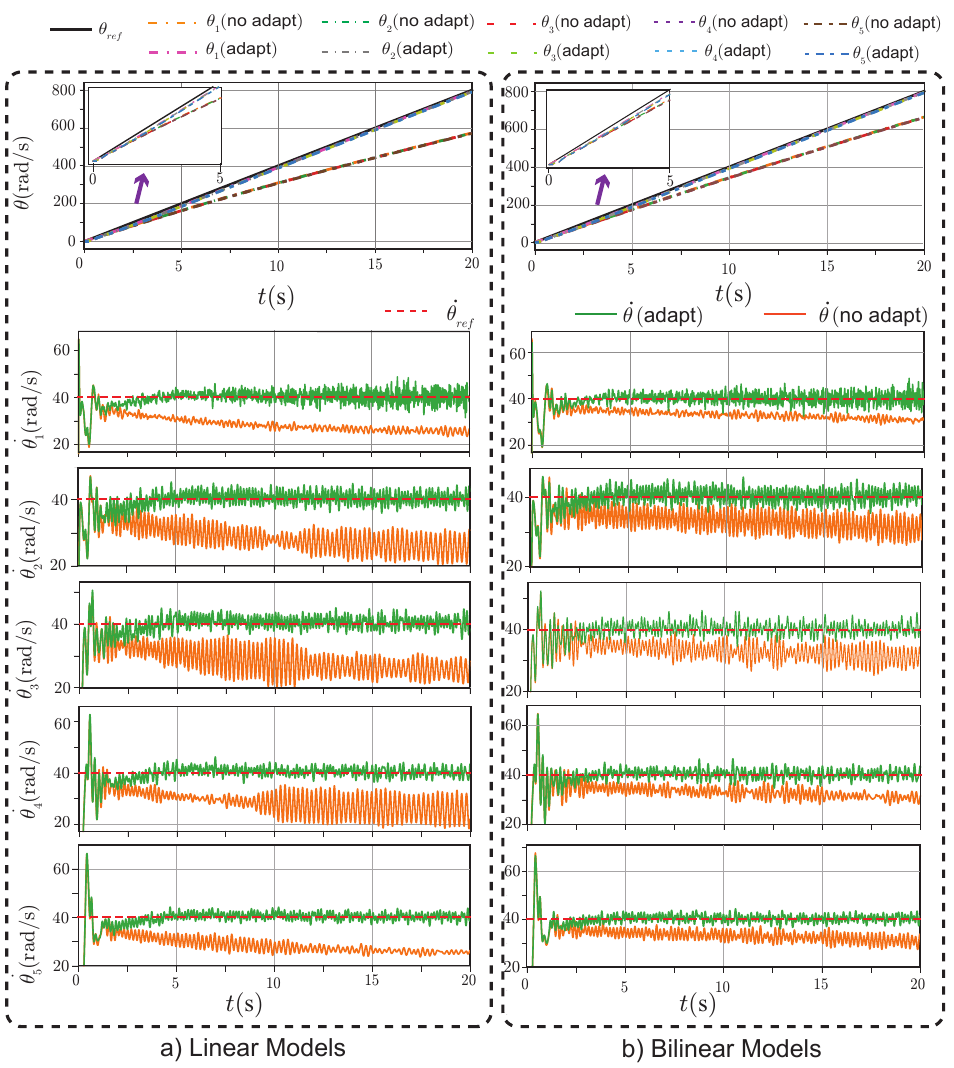}
     \caption{\textbf{Simulation results of the proposed linear(left)/bilinear(right) adaptive Koopman framework for the controlled synchronous revolution of the coupled pendulum system with $\boldsymbol{N=5}$}. The figure presents tracking performance with the nominal system dynamics modified by $\delta = 40 \%$ change in system parameters, and the measured state, $\boldsymbol{x}_{obs}$, is corrupted with the Gaussian noise with SNR = $30$ dB. The adaptive Koopman algorithm is integrated with MPC to drive the coupled pendulum system to track a reference angular velocity $\boldsymbol{\dot{\theta}} = 40$ rad/s. a) Evolution of $\boldsymbol{\theta}$ and $\boldsymbol{\dot{\theta}}$ for the nominal linear Koopman and the linear adaptive Koopman algorithm. b) Evolution of $\boldsymbol{\theta}$ and $\boldsymbol{\dot{\theta}}$ for the nominal bilinear Koopman and the bilinear adaptive Koopman algorithm.}
    \label{fig:coupled_pend}
\end{figure*}

Fig.~\ref{fig:coupled_pend} illustrates the reference and the traced trajectories of $\boldsymbol{\theta}$ and $\boldsymbol{\dot{\theta}}$ for the coupled synchronization problem for $\delta = 40$ and SNR of $30$ dB. The comparison encompasses both linear and bilinear Koopman models, with and without adaptation. Initially, both approaches rely on the same pre-trained nominal Koopman model, leading to similar tracking behavior in the early phase. However, after approximately $1.25$ seconds, discernible divergence occurs. Notably, the adaptive algorithm achieves successful convergence to the reference trajectory in around $3$ seconds, while the non-adaptive algorithm fails to do so. This failure manifests as a constant drift in angle tracking observed in non-adaptive schemes, contrasting with the accurate tracking exhibited by the adaptive scheme. Similar trends are observed for angular velocities, with non-adaptive trajectories diverging and oscillating around a constant value, while adaptive trajectories converge to and oscillate around the reference value. This discrepancy arises because the control signal computed by MPC using nominal Koopman models becomes erroneous due to changes in system dynamics. In contrast, the adaptive algorithm updates the nominal Koopman model online to account for these uncertainties, resulting in precise tracking performances. Additionally, including richer basis functions in the bilinear model contributes to marginally better performance than its linear counterpart, as reflected in the presented results.\par

\begin{figure*}[ht!]
     \centering
     \includegraphics[width=0.98\textwidth]{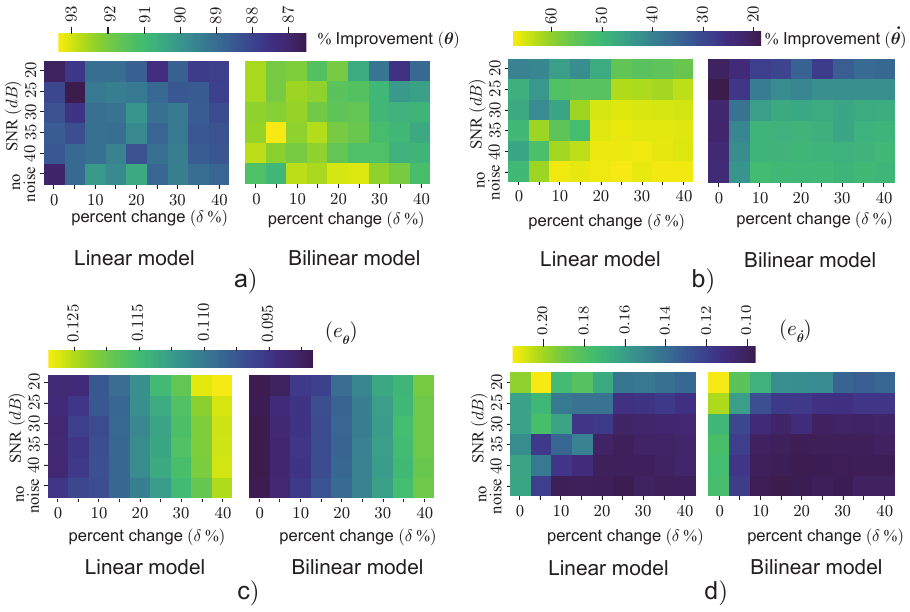}
         \caption{\textbf{Robustness analysis of the proposed adaptive Koopman framework against parametric uncertainty of range $\boldsymbol{\delta \in \{0,5,10,15,20,25,30,35,40}\}$ and measured states $\boldsymbol{x}_{obs}$ corrupted by Gaussian noise with SNR in range $\{\text{no noise}\boldsymbol{, 40, 35, 30, 25, 20}\}$ dB for solving controlled synchronization problem with $\boldsymbol{\dot{\theta}}_{ref} = \boldsymbol{40}$ rad/s}. a) Percent decrease in the average RMS tracking error in $\boldsymbol{\theta}$ for (left: linear, right: bilinear) adaptive Koopman framework compared to nominal Koopman framework. b) Percent decrease in the average RMS tracking error in $\boldsymbol{\dot{\theta}}$ for (left: linear, right: bilinear) adaptive Koopman framework compared to nominal Koopman framework. c) Average RMS tracking error $(e_{\boldsymbol{\theta}})$ in $\boldsymbol{\theta}$ for (left: linear, right: bilinear) adaptive Koopman framework. d) Average RMS tracking error $e_{\boldsymbol{\dot{\theta}}}$ in $\boldsymbol{\dot{\theta}}$ for (left: linear, right: bilinear) adaptive Koopman framework.}
    \label{fig:coup_heat_map_parametric}
\end{figure*}

Figs.~\ref{fig:coup_heat_map_parametric}a and ~\ref{fig:coup_heat_map_parametric}b depict the average reduction in root mean square (RMS) tracking error for the linear and bilinear adaptive Koopman control schemes compared to their non-adaptive counterparts. This heatmap visualization summarizes the results of simulations conducted across a range of parametric uncertainties $(\delta \in  \{5, 10, 15, 20, 25, 30, 35, 40\})$ and measurement noise (SNR $\in$ $\{40, 35, 30, 25, 20\}$ dB), including the scenario with no sensor noise. The heatmap visually demonstrates the substantial reduction in tracking error achieved by the adaptive algorithm over the non-adaptive approach. The error reduction observed in $\boldsymbol{\theta}$ ranges approximately from $87 \%$ to $93\%$. 
Similarly, tracking errors in $\boldsymbol{\dot{\theta}}$ exhibits a minimum reduction of at least $20 \%$, potentially reaching improvements exceeding $60\%$. Figs.~\ref{fig:coup_heat_map_parametric}c and ~\ref{fig:coup_heat_map_parametric}d presents the corresponding heatmap for the same range of conditions for the average tracking errors in $\boldsymbol{\theta}$ and $\boldsymbol{\dot{\theta}}$ for the linear and bilinear adaptive algorithms. The average tracking error for $\boldsymbol{\theta}$ is defined as $e_{{\theta}} =(1/N) \sum_{i=1}^N |{\theta_{i,ref}} - {\theta_i}|/|{\theta_{i,ref}}|$ where $\boldsymbol{\theta}_{ref}$ and $\boldsymbol{\theta}$ are the reference and traced joint angles, respectively. The average tracking error for joint angular velocity $e_{{\dot{\theta}}}$ is defined analogously.\par
Analyzing the scenario with zero parametric uncertainties $(\delta = 0)$ and no sensor noise unveils that even when the system dynamics remain unchanged, non-adaptive Koopman controllers exhibit inferior performance compared to their adaptive counterparts. This exposes a crucial limitation: pre-trained Koopman models, even under ideal conditions, may not perfectly capture true system dynamics. The nominal Koopman neural network might be insufficient for accurately representing a complex system like the coupled pendulum with $N=5$. This might result from insufficient training data, or the offline training process might have identified a sub-optimal model
potentially causing discrepancies between the learned model and the actual system dynamics. The adaptive algorithm achieves significantly improved tracking performance even with potentially suboptimal pre-trained models. Consequently, the adaptive Koopman framework presents a data-efficient approach for learning and control tasks within the Koopman operator-theoretic paradigm. Further, while a nominal Koopman model might offer accurate state predictions, using it directly in closed-loop control can be problematic due to modeling errors in the embedding process (as shown in~\cite{uchida2023control}). These errors can manifest differently during control execution compared to prediction, leading to performance degradation. The proposed adaptive framework addresses this by incorporating an online adaptive module that regularizes the nominal model for the specific control task, similar to the offline network approach in~\cite{uchida2023control}, leading to improved closed-loop control performance. These observations highlight that the adaptive algorithm alleviates the burden of requiring high-fidelity Koopman models, making it suitable for real-world applications. \par

\begin{figure}[ht!]
     \centering
     \includegraphics[width=0.45\textwidth]{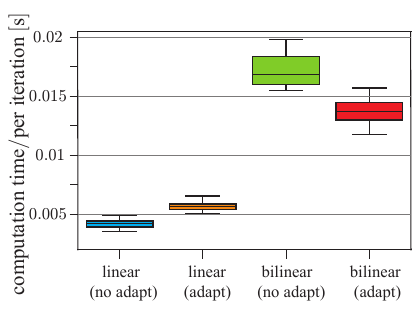}
     \caption{\textbf{Comparison of average computation time per iteration between linear and bilinear nominal Koopman framework with their corresponding adaptive Koopman formulation for solving controlled synchronization problem.} The box plot is generated from simulation data using $42$ different trials with different values of $\delta$ and noise levels.}
    \label{fig:comp_time}
\end{figure}

Figs.~\ref{fig:coupled_pend} and~\ref{fig:coup_heat_map_parametric} reveal comparable tracking performance between the linear and bilinear adaptive algorithms. While the bilinear model exhibits a slight advantage in tracking accuracy, this benefit comes at a high computational cost, as seen in Fig.~\ref{fig:comp_time}. Notably, the linear MPC problem is convex, guaranteeing a unique global optimum that can be efficiently computed. Conversely, the bilinear MPC problem is non-convex, necessitating initialization for the OSQP solver. This solver may struggle to locate the global minimum, leading to increased computation time per iteration in the bilinear case. Since the adaptive algorithm operates online, its computational cost must remain lower than the sampling period for deployment. Although the bilinear approach offers marginally better tracking performance, its substantial computational burden renders it impractical for online applications. Consequently, the remainder of this work focuses solely on the results obtained with the linear adaptive algorithm, prioritizing real-time feasibility over slight performance gains. \par
 
Intriguingly, the non-adaptive bilinear algorithm exhibits a higher computational burden than its adaptive counterpart. This might seem counterintuitive, but it likely stems from the substantial deviations between predicted and desired states in the nominal model necessitating a greater number of iterations within the iterative ADMM (alternating direction method of multipliers) algorithm, employed by OSQP~\cite{osqp}, to achieve convergence for each initialization.
The algorithm iteratively refines the solution by linearizing the bilinear dynamics around the new solution at each step, which iteratively continues until convergence.

\begin{figure*}[ht!]
     \centering
     \includegraphics[width=0.98\textwidth]{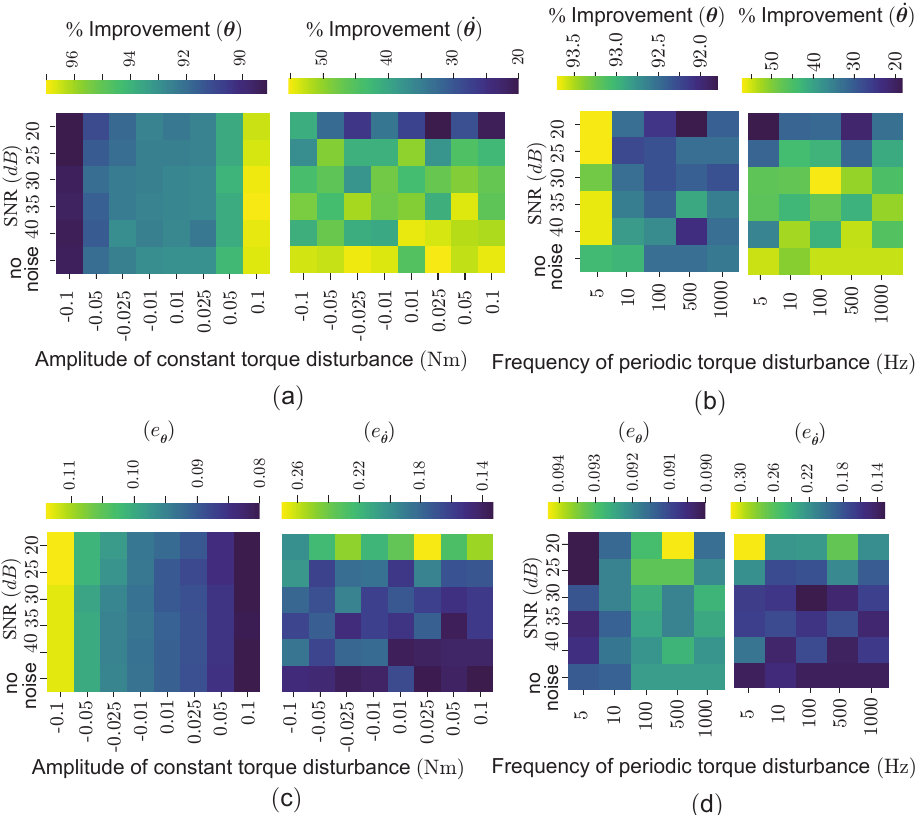}
     \caption{\textbf{Robustness study of the proposed linear adaptive Koopman framework against non-parametric input disturbances and sensor noise}. In particular, two types of input disturbances are considered- constant torque disturbance and time-varying sinusoidal torque disturbance. The constant torque disturbances, denoted as $\tau_{dist}(t) = c$, where $c \in \{-0.1,-0.05, -0.025, -0.01, 0.01, 0.025, 0.05, 0.1\}$ Nm and the time-varying sinusoidal disturbance is written as $\tau_{dist} \in c' \sin(2 \pi f t)$, where $c' = 0.1$ Nm and $f = \{5,10,100,500,1000\}$ Hz. a) Percentage decrease in average RMS tracking error in $\boldsymbol{\theta}$ (left) and $\boldsymbol{\dot{\theta}}$ (right) for linear adaptive Koopman algorithm as compared to the nominal linear Koopman framework for constant input disturbance. b) Percentage decrease in average RMS tracking error in $\boldsymbol{\theta}$ (left) and $\boldsymbol{\dot{\theta}}$ (right) for linear adaptive Koopman algorithm as compared to the nominal linear Koopman framework for constant sinusoidal input disturbance. c) Average RMS tracking error in $\boldsymbol{\theta}$ (left) and $\boldsymbol{\dot{\theta}}$ (right) for linear adaptive Koopman algorithm for constant input torque disturbances. d) Average RMS tracking error in $\boldsymbol{\theta}$ (left) and $\boldsymbol{\dot{\theta}}$ (right) for linear adaptive Koopman algorithm for sinusoidal input torque disturbances.}
    \label{fig:coup_heat_map_nonparametric}
\end{figure*}

\textbf{\emph{Adaptation to Non-parametric Uncertainties: }}
Now, we evaluate the performance of the adaptive linear Koopman control algorithm under non-parametric uncertainties. In particular, we explore two types of non-parametric uncertainties: a constant input torque disturbance and a time-varying sinusoidal input torque disturbance. Additionally, sensor noise with the same amplitude range as the previous case is introduced. It is crucial to demonstrate that the effectiveness and robustness of the adaptive algorithm observed in parametric settings generalize to non-parametric scenarios as well because for parametric uncertainties, unlike their non-parametric counterparts, it cannot be guaranteed that the system dynamics still resides within the invariant Koopman subspace spanned by pre-learned observables.

Fig.~\ref{fig:coup_heat_map_nonparametric} presents simulation results depicting the performance of the adaptive linear algorithm in the presence of non-parametric input disturbances and sensor noise. The constant non-parametric input torque disturbance and the time-varying sinusoidal input disturbance are mathematically expressed as $\tau_{dist} = c$, where $c \in \{-0.1, -0.05, -0.025, -0.01, 0.01, 0.025, 0.05, 0.1\}$ Nm and $\tau_{dist}(t) = c' \sin(2 \pi f t)$, where $f \in \{5, 10, 100, 500, 1000\}$ Hz is the frequency and $c' = 0.1$ Nm is the disturbance amplitude. Figs.~\ref{fig:coup_heat_map_nonparametric}a and~\ref{fig:coup_heat_map_nonparametric}b depict the improvement in the average RMS tracking error for constant and sinusoidal disturbances, respectively. For the constant torque disturbance case, the reduction in tracking error for $\boldsymbol{\theta}$ ranges  $89.1 \%$ to $97 \% $, while for $\boldsymbol{\dot{\theta}}$, it ranges from $20 \%$ to  $55.3 \%$. This reduction is observed across all tested amplitudes and SNR levels, confirming the effectiveness of the adaptive algorithm in adapting to non-parametric disturbances. Similar results are obtained for the sinusoidal disturbance case across the chosen frequency range and SNR. Figs.~\ref{fig:coup_heat_map_nonparametric}c and ~\ref{fig:coup_heat_map_nonparametric}d present the corresponding average tracking errors of the linear adaptive Koopman algorithm for constant and sinusoidal torque disturbances, respectively. The consistently low average errors for both $\boldsymbol{\theta}$ and $\boldsymbol{\dot{\theta}}$ demonstrate the effectiveness of the adaptive algorithm in achieving the control objective, even in the presence of non-parametric disturbances. 

These observations show that the efficacy of the adaptive algorithm extends to scenarios encompassing uncertainties that may not fully lie in the subspace spanned by the pre-learned basis function. This is because when sufficiently rich basis functions are learned during offline training, the uncertainties can be approximated with adequate accuracy by projecting them onto these basis functions. This enables the adaptive algorithm to learn these uncertainties online by adapting the existing $\boldsymbol{A}$ and $\boldsymbol{B}$ matrices without the need to retrain the lifting function $\boldsymbol{\varphi}(\cdot)$.

\subsubsection{Serial Manipulator} \label{sec:Manipulator}
\begin{figure*}[ht!]
     \centering
     \includegraphics[width=\textwidth]{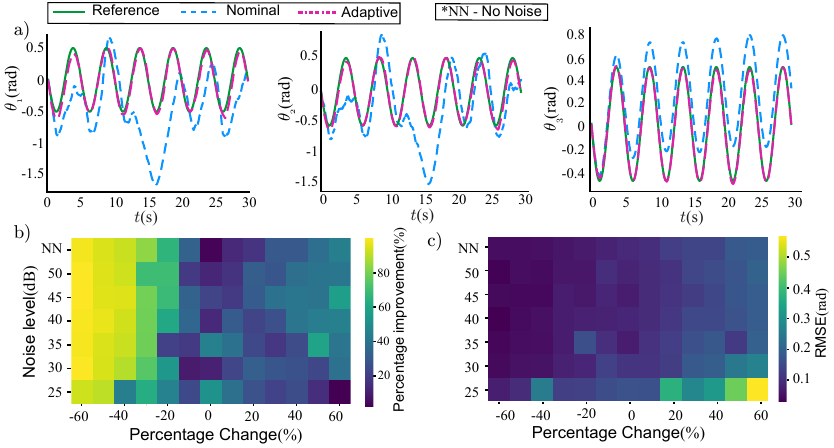}
     \caption{\textbf{Simulation results for the 3R serial manipulator}. Shown are the results for tracking control of a 3R serial manipulator for changes in mass (-60\% to 60\%) and inclusion of noise (25$-$50 dB). a) Evolution of joint angles with mass change of -50\% and noise level of 25 dB. b) Percentage improvement in tracking error for the adaptive linear Koopman models over the nominal linear model. c) Joint position RMSE error for the adaptive linear Koopman model.}
    \label{fig:manip_res}
\end{figure*}
As our second example, we consider trajectory tracking for a 3R serial manipulator. Serial robots, with their multiple interconnected links, exhibit intricate dynamic behaviors, which, along with their widespread application, makes them a prime candidate to benchmark the performance of the proposed algorithm. 

The system dynamics for a serial manipulator~\cite{ogata2010modern} is given as:
\begin{eqnarray}
    \label{eq:manip_dyn}
    \boldsymbol{M}(\boldsymbol{\theta})\boldsymbol{\ddot{\theta}} + \boldsymbol{C}(\boldsymbol{\theta}, \,\boldsymbol{\dot{\theta}}) + \boldsymbol{G}(\boldsymbol{\theta}) = \boldsymbol{\tau},
\end{eqnarray}
where $\boldsymbol{\theta} \in \mathbb{R}^3$ and $\boldsymbol{\tau} \in \mathbb{R}^3$ represent the joint angles and the torque inputs, respectively. $\boldsymbol{M}(\cdot) \in \mathbb{R}^{3 \times 3}$, $\boldsymbol{C}(\cdot, \cdot) \in \mathbb{R}^{3 \times 3}$ and $\boldsymbol{G}(\cdot) \in \mathbb{R}^{3\times 1}$ represent the mass, the Coriolis and the gravity matrix, respectively. To train the nominal linear model, we take the system parameters of each link to be $m_i = 0.8 \;\text{kg}, l_i = 1\;\text{m}, I_i = \frac{m_{i}l_{i}^{2}}{12}\;\text{kgm$^2$}, $ for $i = 1,2,3$, where $m_i, I_i$ and $l_i$ represent the mass, inertia, and length of $i^{th}$ link, respectively. 

In order to conduct a comprehensive analysis of the efficacy of the proposed linear adaptive Koopman algorithm, we have performed a robustness study with varying levels of mass variations to the links along with the addition of noise to the measured states $\boldsymbol{x}_{obs} = [\boldsymbol{\theta}^{\top},\boldsymbol{\dot{\theta}}^{\top}]^{\top}$. The considered variation for the mass of all the links ranges from $-60\%$ to $60\%$. Meanwhile, the SNR values for the noise range from $50$dB to $25$dB. For the current study, we illustrate trajectory tracking control, wherein feedback control is implemented to ensure all the joints follow a reference sinusoidal path. The tracking results for a particular case where we reduce the mass of each link, and consequently its inertia, by $50\%$ along with the inclusion of measurement noise of $25$dB shown in Fig.~\ref{fig:manip_res}a. Similar to the prior example, during the initial phase lasting approximately one second, both the nominal and adaptive control architectures exhibit similar behavior in tracking the reference trajectory. However, beyond that, notable deviations are observed in the traced joint angles as compared to the specified reference sinusoidal signal for the nominal model. It can be seen that the tracking performance of the nominal model is significantly worse for lower joints (${\theta_1}$ and ${\theta_2}$) than the higher joint (${\theta_3}$). This can be explained by the fact that the torque required to actuate a joint is heavily dependent on the links that follow it. As such, the torque required by the first joint to trace the desired trajectory is affected by the mass change of all the three links that follow it, while the torque required by the third joint is only affected by the mass change of the last link. Despite these complexities, the adaptive Koopman algorithm demonstrates exceedingly better tracking performance for all three joints. The corresponding root mean square tracking errors for the adaptive and nominal models are 0.189 rad and 1.27 rad, respectively, corresponding to a $92\%$ improvement in the tracking error. A visual analysis of the performance of the adaptive Koopman algorithm over the complete range of induced variations in the form of heatmaps is shown in Figs.~\ref{fig:manip_res}b and~\ref{fig:manip_res}c. It can be seen that the adaptive Koopman model consistently outperforms the nominal Koopman model with a percentage improvement in tracking error up to $90\%$. 


\subsubsection{Planar Quadrotor}\label{sec:plan_quad}
\begin{figure*}[ht!]
     \centering
     \includegraphics[width=\textwidth]{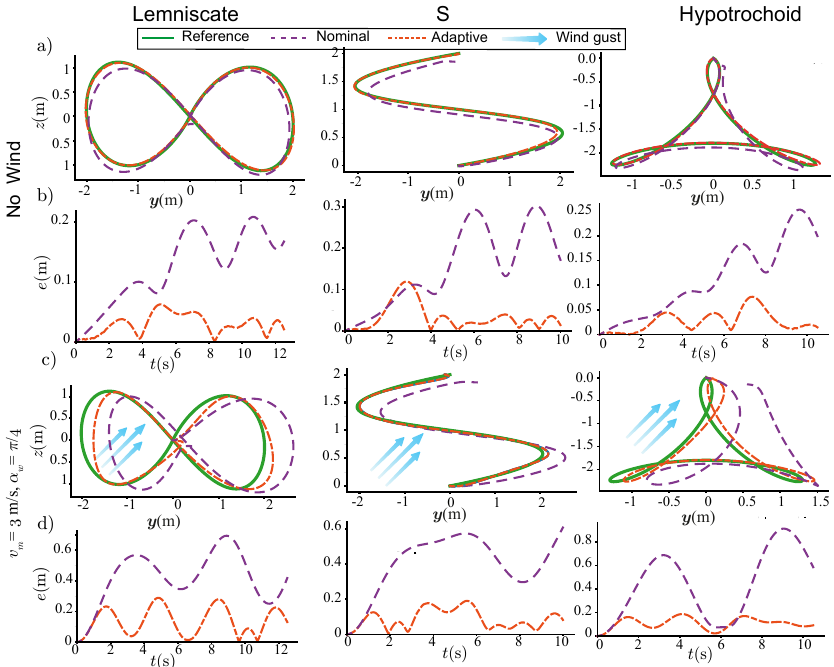}
     \caption{\textbf{Simulation results for the planar quadrotor.} Shown are the results for tracking control of a planar quadrotor whose mass has been changed by 40\% post training of the nominal model in the presence (c,d) and absence of wind gusts (a,b) for different shapes (Lemniscate, S-shape, and Hypotrochoid). The 
 wind gust has a mean velocity of 3 m/s with a direction of 45$^\circ$ anticlockwise w.r.t to the $y$-coordinate axis. a,c) Paths traced. b,d) Evolution of tracking errors.}
    \label{fig:plana_quad}
\end{figure*}
To conclude our exploration of the algorithm's capabilities, we evaluate the adaptive linear Koopman algorithm for tracking control of a planar quadrotor. This choice serves a two-fold purpose. Firstly, it showcases the algorithm's effectiveness on a mobile robot platform, whose dynamics is inherently different compared to the previously studied systems. This application highlights the algorithm's ability to adapt to diverse robotic systems. Secondly, we further emphasize the algorithm's generality and platform-agnostic nature by successfully controlling a system with agile dynamics. 

For this simulation, in addition to parametric variations, we also incorporate wind gusts in the simulation environment to show the robustness of the proposed scheme against changes in system dynamics induced by environmental factors, which are inherently non-parametric. The system dynamics for a planar quadrotor in the presence of wind gusts is given by~\cite{folkestad2021diction}:

\begin{align}
    \label{eq:planar_quad_dynamics}
    &\begin{bmatrix}
        \ddot{y} \\ \ddot{z} \\ \ddot{\theta} 
    \end{bmatrix} {=}     \begin{bmatrix}
        0 \\ -g \\ 0 
    \end{bmatrix} {+}  \begin{bmatrix}
        -\frac{1}{m} sin{\theta} &&  -\frac{1}{m}sin{\theta} \\
         \frac{1}{m} cos{\theta} &&  \frac{1}{m}cos{\theta} \\ 
        -\frac{l_{arm}}{I} && \frac{l_{arm}}{I}
    \end{bmatrix} \begin{bmatrix}
        T_1 \\ 
        T_2 
    \end{bmatrix} + \frac{\boldsymbol{F}_w}{m}, \nonumber \\
    &\boldsymbol{F}_w = \begin{bmatrix}
        K v_w^{2}cos(\alpha_w),
        K v_w^{2}sin(\alpha_w),
        0
    \end{bmatrix}^\top,
\end{align}

where $y, z$ and $\theta$ represent the Cartesian position and orientation in the $y-z$ plane, respectively. $T_1$ and $T_2$ are the thrust inputs for the propellers. $m,\, l_{arm},\, I$, and $g$ refer to the mass, length of the propeller arm, rotational inertia, and gravity, respectively. $\boldsymbol{F}_w$ represents the drag force exerted by the wind on the planar quadrotor. $v_w$ and $\alpha_w$ represent the speed and direction of the wind gust, respectively. $K$ is the drag force coefficient. Initially, the parameters are taken as $m = 2\,\text{ kg}$, $I = 1\,\text{ kgm/s}^{2}$, $g = 9.81\,\text{ m/s}^{2}$, $l_\text{arm} = 0.2\,\text{ m}$, $K = 0.1$ \text{kg/m},$v_w = 0 \text{ m/s}$ and $\alpha_w = 0  \text{ rad}$ to generate the input-output dataset for training the nominal linear Koopman model. For this system, we perform trajectory tracking across multiple shapes, as shown in Fig.~{\ref{fig:plana_quad}}. During deployment, we change the planar quadrotor parameters to $m = 2.4\,\text{kg}$, $I = 1.2\,\text{kgm/s}^{2}$. The corresponding results are shown in Figs.~{\ref{fig:plana_quad}}a and \ref{fig:plana_quad}b. It can be seen that the adaptive Koopman architecture offers significantly better tracking performance than the nominal model for all the chosen paths. Subsequently, we also introduce wind gusts with velocity $v_w = 3 \text{m/s}$ and direction $\alpha_w = \frac{\pi}{4} \text{rad}$. The corresponding results are shown in Figs.~{\ref{fig:plana_quad}}c and \ref{fig:plana_quad}d. Similar to the previous case, the adaptive architecture significantly outperforms the nominal model for tracking the reference paths.

Hence, the presented simulation studies demonstrate the efficacy of the proposed algorithm in adapting to changes in system dynamics resulting from both intrinsic and environmental factors.

\section{Discussion}
\label{sec:Conc}
This work presents an adaptive Koopman formulation for the control of nonlinear systems, capable of adapting online to disturbances, noise, and variations in system dynamics. The framework leverages a pre-trained Koopman model alongside an online adaptive module for real-time adaptation to the aforementioned variations. This online learning capability empowers the framework to achieve robust control performance under uncertainties. The proposed framework demonstrates efficacy through the successful implementation of tracking control within an MPC framework across diverse robotic paradigms, including coupled pendulums, serial manipulators, and planar quadrotors. This highlights the framework's generalizability and robustness across various robotic systems. The robustness is further emphasized by exploring the performance of the adaptive algorithm for these systems to a wide range of parametric and non-parametric uncertainties, measurement noise, input disturbances, and even environmental changes (e.g., wind gusts). This comprehensive evaluation underscores the superior performance of the adaptive architecture compared to the nominal Koopman model. Notably, the algorithm demonstrates robustness while adapting to non-parametric uncertainties that may not strictly satisfy the underlying theoretical assumptions behind the adaptive algorithm. 

The efficacy of the proposed adaptive algorithm in achieving robust control performance across diverse, complex systems with varied uncertainties and limited online data for adaptation presents a compelling contrast to the offline model's requirement for a significantly larger dataset for accurate model training. This disparity can be attributed to the fundamental difference in their learning approaches. Unlike the offline model, which aims to learn a globally valid Koopman representation, the adaptive algorithm strategically tailors corrections to the nominal Koopman model only in the local vicinity of the current state. This targeted adaptation facilitates accurate control performance within the local operating region while circumventing the need for extensive offline data collection.

Beyond its robustness, the proposed adaptive Koopman framework offers a significant improvement in achieving control objectives even with inaccurate Koopman models. Traditionally, generating highly accurate Koopman models of complex robotic systems from data is a challenging and resource-intensive endeavor. This framework, however, overcomes this limitation by effectively leveraging faulty Koopman models for high-fidelity control performance. This ability to compensate for initial model inaccuracies paves the way for a broader application of Koopman operator theory in practical robotic control tasks. Furthermore, the proposed adaptive algorithm offers a compelling advantage by enabling the utilization of computationally efficient linear Koopman models. Typically, achieving high accuracy with traditional linear Koopman models for complex systems necessitates a large number of observable functions, significantly increasing computational cost. Consequently, researchers often resort to computationally expensive bilinear models. However, these bilinear models retain a degree of nonlinearity, limiting the applicability of established linear control techniques. The proposed framework overcomes this trade-off by empowering the use of simpler, computationally efficient linear Koopman models while achieving prediction accuracy comparable to bilinear models through online adaptation. This reduces the computational burden and unlocks the vast toolbox of well-developed linear control methods for complex nonlinear systems.

Despite the numerous advantages, the requirement for manual hyperparameter tuning (window size, epochs, regularization parameters) of the online adaptive module remains a limitation of this study. While simulations indicate robust performance within a specific hyperparameter range, this range needs manual discovery. Thus, integrating algorithms for automated hyperparameter tuning within the online architecture is a major focus of future work. Additionally, while simulations demonstrate robustness under significant parametric variations and noise, experimental validation with real robots exhibiting large parameter changes presents a fascinating and challenging as the OSQP solver, employed to solve the MPC optimization problem, may encounter difficulty converging in these scenarios, compromising the ability of the adaptive scheme to handle extremely large uncertainties. Our future endeavors aim to translate these promising simulation results into quantitative performance metrics through controlled experiments on a diverse range of physical systems. This will further solidify the potential of the adaptive Koopman framework for expanding the applicability of Koopman theory to real-world robotic control.  

\section{Methods}

 \subsection{Koopman Operator Theory for Control-Affine Systems}
 \label{sec:Koop_cont_affine}
 The time evolution of the the time-varying observable, $\boldsymbol{\psi}(t, \boldsymbol{x})$, for the control-affine system~(\ref{eq:base_affine}) is given by \cite{goswami2017global}
\begin{eqnarray}
\label{eq:eig_psi_affine}
    \frac{\partial{\psi}}{\partial{t}} = \mathcal{L}_{\boldsymbol{f_0}} \psi + \sum_{i=1}^{m} u_i \mathcal{L}_{\boldsymbol{f}_i} \psi, \;\;\psi(0,\boldsymbol{x}) = \vartheta(\boldsymbol{x_0}),
\end{eqnarray}

where $\mathcal{L}_{\boldsymbol{f}_i}= \boldsymbol{f}_i\cdot\nabla \, \forall \, i {=} 1,...,m$ are the corresponding Lie derivatives. Often, KEFs are the preferred choice of the basis function for the observable space. The mathematical transformation of the state space basis to KEFs is known as the Koopman Canonical Transform (KCT) \cite{surana2016koopman}.
Given a sufficiently large number of basis functions, any observable function can be expressed in terms of KEFs within the observable space. Therefore, the state vector $ \boldsymbol{x}$ can be written as $\boldsymbol{x} = \sum_{i=1}^{p} \phi_i(\boldsymbol{x}) \boldsymbol{v}_i^{\boldsymbol{x}}, \, p \to \infty$; where $\phi_i(\cdot)$ represents the KEF and $\boldsymbol{v}_i^{\boldsymbol{x}} \in \mathbb{C}^p$ represents the corresponding Koopman eigenmodes (KEMs). Invoking the transformation $\boldsymbol{z} = \boldsymbol{T}(\boldsymbol{x}) = [{\phi}_1(\boldsymbol{x}),{\phi}_2(\boldsymbol{x}),...,{\phi}_{p}(\boldsymbol{x})]^{T}$, the system (\ref{eq:base_affine}) can be represented in terms of the Koopman eigen-coordinates as:
\begin{eqnarray}
    \label{eq:kct}
    \boldsymbol{\dot{z}} {=} \boldsymbol{\mathcal{L}_{f_0} z} + \sum_{i=1}^{m} \mathcal{L}_{\boldsymbol{f}_i} \boldsymbol{T} \;(\boldsymbol{x})u_i|_{\boldsymbol{x=C z}},\;
    \boldsymbol{x} {=} \boldsymbol{C z},
\end{eqnarray}

where $\boldsymbol{C} = [\boldsymbol{{v}_1^{\boldsymbol{x}}}|...|\boldsymbol{v}_n^{\boldsymbol{x}}]$. 

The dynamics expressed in canonical coordinates (\ref{eq:kct}) remain nonlinear with respect to the control term. In order to linearize/bilinearize the Koopman dynamics, the following assumptions are considered:

\textit{Assumption - 1  - There exist a finite set of KEFs of the drift vector field, $f_00$ such that $\{\phi_j : j=1,...,p \}, p \in \mathbb{N}, p < \infty$ and the span$\{\phi_1,..., \phi_p \}$ forms an invariant subspace of $\mathcal{L}_{\boldsymbol{f}_0}$, i.e., $\mathcal{L}_{\boldsymbol{{f}}_0} \boldsymbol{T}(\boldsymbol{x}) \in \{ \phi_1,..., \phi_p \}$}.

\textit{Assumption - 2  - The span$\{\phi_1,..., \phi_p \}$ forms an invariant subspace of $\mathcal{L}_{f_i}, \forall i=1,...,m$}.

\textit{Assumption - 3 - $\mathcal{L}_{\boldsymbol{{f}}_i} \boldsymbol{T}(\boldsymbol{x}) \boldsymbol{u} \in span\{u_1, u_2, ..., u_m\}$}.

Under assumptions 1-2, the continuous-time finite dimensional Koopman bilinear dynamics for system~(\ref{eq:base_affine}) can be obtained as
\begin{eqnarray}
    \boldsymbol{\dot{z}} = \boldsymbol{A_c} \boldsymbol{z} + \boldsymbol{B}_{\boldsymbol{c}} (\boldsymbol{z} \otimes \boldsymbol{u}), \boldsymbol{x} = \boldsymbol{C}\boldsymbol{z}. \nonumber
\end{eqnarray}
A detailed proof can be found in~\cite{goswami2017global}. Further, given assumption 3, a linear Koopman representation can similarly be obtained as
\begin{eqnarray}
\dot{\boldsymbol{z}} = \boldsymbol{A}_{c}\boldsymbol{{z}}{+} \boldsymbol{B}_{c}\boldsymbol{u},  \boldsymbol{x} = \boldsymbol{C}\boldsymbol{z}. \nonumber
\end{eqnarray}
\emph{Remark 1:} Assumptions 1-3 require the span of KEFs to be an invariant subspace under the action of the Koopman operator $\mathcal{K}$, which may seem restrictive in the context of practical implementation where such closure guarantees may not necessarily hold. Nevertheless, extensive research has demonstrated that finite-dimensional linear~\cite{uchida2023control, zhang2022robust} and bilinear models~\cite{sah2024real, bruder2021advantages, folkestad2022koopnet} can be synthesized with commendable accuracy by invoking these assumptions for a wide variety of nonlinear dynamical systems. 

The corresponding discrete-time representations for the linear/bilinear Koopman dynamics can be given as

\begin{align}
    &\boldsymbol{{z}}_{k{+}1} {=} \boldsymbol{A}\boldsymbol{z}_k {+} \boldsymbol{B}\boldsymbol{u}_k, \,\, \boldsymbol{{x}}_{k+1} {=} \boldsymbol{C}\boldsymbol{{z}}_{k+1}, \\
    &\boldsymbol{{z}}_{k{+}1} {=} \boldsymbol{A}\boldsymbol{z}_k {+} \boldsymbol{B}(\boldsymbol{z}_k\otimes\boldsymbol{u}_k), \,\, \boldsymbol{{x}}_{k+1} {=} \boldsymbol{C}\boldsymbol{{z}}_{k+1},
\end{align}
where $\boldsymbol{{z}}_{k{+}1}$ and $\boldsymbol{{x}}_{k+1}$ are the one-step ahead lifted state and base state, respectively.

\subsection{Theoretical Analysis}
\label{sec:adaptation_theory}
Under conditions similar to assumptions 1-3, we now undertake a formal synthesis of continuous-time adaptive bilinear/linear Koopman models by invoking the following theorems.

\textit{Theorem 1 - Under the assumption $\mathcal{L}_{\boldsymbol{\tilde{f}}_0} \boldsymbol{T}(\boldsymbol{x}),\,\mathcal{L}_{\boldsymbol{\tilde{f}}_i} \boldsymbol{T}(\boldsymbol{x}) \in span\{ \phi_1(\boldsymbol{x}),{\phi}_2(\boldsymbol{x}),...,{\phi}_{p}(\boldsymbol{x}) \}$, the uncertain control-affine system (\ref{eq:affine_mod}) can be compensated via adapted Koopman matrices $\boldsymbol{\Delta A}_c$ and $\boldsymbol{\Delta B}_c$ within the bilinear framework~(\ref{eq:adap_bil}), where $\boldsymbol{\Delta A}_c$ and $\boldsymbol{\Delta B}_c$ are the continuous-time analogues of the adaptive matrices $\boldsymbol{\Delta A}$ and $\boldsymbol{\Delta B}$ in~(\ref{eq:adap_bil}).}

\emph{Proof:} Invoking the KCT transformation, the system (\ref{eq:affine_mod}) in the new coordinates becomes:

\begin{align}
\dot{\boldsymbol{z}} =& \frac{\partial\boldsymbol{T}(\boldsymbol{x})}{\partial\boldsymbol{x}}\dot{\boldsymbol{x}}  
 \nonumber \\
= &\frac{\partial\boldsymbol{T}(\boldsymbol{x})}{\partial\boldsymbol{x}} \left( \boldsymbol{f_0}(\boldsymbol{x}) + \boldsymbol{\tilde{f}_0}(\boldsymbol{x})+ \sum_{i=1}^{m} (\boldsymbol{f}_i(\boldsymbol{x}) + \boldsymbol{\tilde{f}}_i(\boldsymbol{x}))u_i\right) \nonumber \\
= &\mathcal{L}_{\boldsymbol{f}_0}\boldsymbol{T} \;(\boldsymbol{x}) {+} \sum_{i=1}^{m} \mathcal{L}_{\boldsymbol{f}_i} \boldsymbol{T} \;(\boldsymbol{x})u_i + \mathcal{L}_{\boldsymbol{\tilde{f}}_0}\boldsymbol{T} \;(\boldsymbol{x}) \nonumber\\ &{+}\sum_{i=1}^{m} \mathcal{L}_{\boldsymbol{\tilde{f}}_i} \boldsymbol{T} \;(\boldsymbol{x})u_i. \nonumber 
\end{align}

Under assumptions 1 and 2, we have
\begin{eqnarray}
\boldsymbol{\dot{z}}= \boldsymbol{A}_c\boldsymbol{{z}}{+} \boldsymbol{B}_c(\boldsymbol{z} 
\otimes \boldsymbol{u}) + \mathcal{L}_{\boldsymbol{\tilde{f}}_0}\boldsymbol{T} \;(\boldsymbol{x}) {+} \sum_{i=1}^{m} \mathcal{L}_{\boldsymbol{\tilde{f}}_i} \boldsymbol{T} \;(\boldsymbol{x})u_i, \nonumber
\end{eqnarray}

Now, if $\mathcal{L}_{\boldsymbol{\tilde{f}}_0} \boldsymbol{T}(\boldsymbol{x}) \in span\{ \phi_1(\boldsymbol{x}),{\phi}_2(\boldsymbol{x}),...,{\phi}_{p}(\boldsymbol{x}) \}$ and $\mathcal{L}_{\boldsymbol{\tilde{f}}_i} \boldsymbol{T}(\boldsymbol{x}) \in span\{ \phi_1(\boldsymbol{x}),{\phi}_2(\boldsymbol{x}),...,{\phi}_{p}(\boldsymbol{x}) \}$, then $\mathcal{L}_{\boldsymbol{\tilde{f}}_0} {z}_{i} = \boldsymbol{\Delta a}_{c,i} {z}_i, \boldsymbol{\Delta a}_{c,i} \in \mathbb{R}^{N}$ and $\mathcal{L}_{\boldsymbol{\tilde{f}}_i} \boldsymbol{z} {u}_{i} = \boldsymbol{\Delta B}_{c,i} \boldsymbol{z} u_i, \boldsymbol{\Delta B}_{c,i} \in \mathbb{R}^{N \times N}$. Hence 
\begin{eqnarray}
\dot{\boldsymbol{z}} &=& \left(\boldsymbol{A}_c + \boldsymbol{\Delta A}_c\right)\boldsymbol{{z}}{+} \boldsymbol{B}_{c}(\boldsymbol{z}\otimes\boldsymbol{u}) +   \sum_{i=1}^{m}\boldsymbol{\Delta B}_{c,i} \boldsymbol{z} u_i,\nonumber \\
&=& \left(\boldsymbol{A}_{c} + \boldsymbol{\Delta A}_{c}\right)\boldsymbol{{z}}{+} \left(\boldsymbol{B}_c +  \boldsymbol{\Delta B}_{c}\right)(\boldsymbol{z}\otimes\boldsymbol{u}) \nonumber
,
\end{eqnarray}
where $\boldsymbol{\Delta A}_c = [\boldsymbol{\Delta a}_{c,1} \; \boldsymbol{\Delta a}_{c,2} \; ... \; \boldsymbol{\Delta a}_{c,N}] \in \mathbb{R}^{N \times N}$ and $\boldsymbol{\Delta B}_c = [\boldsymbol{\Delta B}_{c,1} \;\boldsymbol{\Delta B}_{c,2} \; ... \; \boldsymbol{\Delta B}_{c,m}] \in \mathbb{R}^{N \times Nm}$.

\textit{Theorem 2 - Under the assumptions $\mathcal{L}_{\boldsymbol{\tilde{f}}_0} \boldsymbol{T}(\boldsymbol{x}) \in span\{ \phi_1(\boldsymbol{x}),{\phi}_2(\boldsymbol{x}),...,{\phi}_{p}(\boldsymbol{x}) \}$ and $\mathcal{L}_{\boldsymbol{{\tilde{f}}}_i} \boldsymbol{T}(\boldsymbol{x}) \boldsymbol{u} \in span\{u_1, u_2, ..., u_m\}$, the uncertain control-affine system (\ref{eq:affine_mod}) can be compensated via adapted Koopman matrices $\boldsymbol{\Delta A}_c$ and $\boldsymbol{\Delta B}_c$ within the linear framework~(\ref{eq:adap_lin}), where $\boldsymbol{\Delta A}_c$ and $\boldsymbol{\Delta B}_c$ are the continuous-time analogues of the adaptive matrices $\boldsymbol{\Delta A}$ and $\boldsymbol{\Delta B}$ in~(\ref{eq:adap_lin}).}

\emph{Proof:} Invoking the KCT transformation, the system (\ref{eq:affine_mod}) in the new coordinates becomes:
\begin{eqnarray}
\dot{\boldsymbol{z}} &=&\mathcal{L}_{\boldsymbol{f}_0}\boldsymbol{T} \;(\boldsymbol{x}) {+} \sum_{i=1}^{m} \mathcal{L}_{\boldsymbol{f}_i} \boldsymbol{T} \;(\boldsymbol{x})u_i + \mathcal{L}_{\boldsymbol{\tilde{f}}_0}\boldsymbol{T} \;(\boldsymbol{x}) \nonumber\\ & &{+}\sum_{i=1}^{m} \mathcal{L}_{\boldsymbol{\tilde{f}}_i} \boldsymbol{T} \;(\boldsymbol{x})u_i. \nonumber
\end{eqnarray}

Under assumptions 1 and 3, we have
\begin{eqnarray}
\boldsymbol{\dot{z}}= \boldsymbol{A}_{c}\boldsymbol{{z}}{+} \boldsymbol{B}_{c}\boldsymbol{u} + \mathcal{L}_{\boldsymbol{\tilde{f}}_0}\boldsymbol{T} \;(\boldsymbol{x}) {+} \sum_{i=1}^{m} \mathcal{L}_{\boldsymbol{\tilde{f}}_i} \boldsymbol{T} \;(\boldsymbol{x})u_i.  \nonumber
\end{eqnarray}

Now, if $\mathcal{L}_{\boldsymbol{\tilde{f}}_0} \boldsymbol{T}(\boldsymbol{x}) \in span\{ \phi_1(\boldsymbol{x}),{\phi}_2(\boldsymbol{x}),...,{\phi}_{p}(\boldsymbol{x}) \}$ and $\mathcal{L}_{\boldsymbol{{\tilde{f}}}_i} \boldsymbol{T}(\boldsymbol{x}) \boldsymbol{u} \in span\{u_1, u_2, ..., u_m\}$, 
then $\mathcal{L}_{\boldsymbol{\tilde{f}}_o} {z}_{i} = \boldsymbol{\Delta a}_{c,i}{z}_i, \boldsymbol{\Delta a}_{c,i} \in \mathbb{R}^{N}$ and $\mathcal{L}_{\boldsymbol{\tilde{f}}_i}\boldsymbol{z}{u}_{i} = \boldsymbol{\Delta b}_{c,i}{u}_i, \boldsymbol{\Delta b}_{c,i} \in \mathbb{R}^{N}$. Hence 
\begin{eqnarray}
\dot{\boldsymbol{z}} = \left(\boldsymbol{A}_c + \boldsymbol{\Delta A}_{c}\right)\boldsymbol{{z}}{+} \left(\boldsymbol{B}_{c} +  \boldsymbol{\Delta B}_{c}\right)\boldsymbol{u}, \nonumber
\end{eqnarray}
where $\boldsymbol{\Delta B}_c = [\boldsymbol{\Delta b}_{c,1}\; \boldsymbol{\Delta b}_{c,2}\; ... \;\boldsymbol{\Delta b}_{c,m}] \in \mathbb{R}^{N \times m}$.


\subsection{Network Architecture}\label{sec:neural_arc}
This investigation explores two adaptive Koopman frameworks: linear and bilinear. Each framework leverages a distinct neural network architecture specifically tailored to the targeted Koopman representation. These architectures share a fundamental structure comprised of two distinct network architectures:
\begin{itemize}
\item Nominal Koopman Neural Network: This network, trained offline on the nominal system dataset, aims to capture a global Koopman representation of the nominal system dynamics. Upon training, the nominal Koopman neural network generates mapping $\boldsymbol{\varphi}(\cdot)$ from state space to observable space, and the Koopman model matrices $\boldsymbol{A}$, $\boldsymbol{B}$ and $\boldsymbol{C}$.

\item Adaptive Neural Network: This neural network learns online to refine the Koopman model in real-time by correcting for deviations between the nominal model and the true system dynamics.
\end{itemize}
Figure~\ref{fig:adaptation_block} provides a schematic illustration of both the linear and bilinear Koopman network architectures, along with their corresponding adaptation networks. Despite their distinct model representation (linear vs. bilinear), both network types share a core foundational architecture. The nominal Koopman architecture for both the bilinear and linear models consists of three key components (refer to~\cite{sah2024real} for more comprehensive details):
\begin{itemize}
\item Lifting block: This block consists of a fully connected deep neural network that learns the mapping $\boldsymbol{\varphi}(.)$ from the base state ($\boldsymbol{x}$) into the lifted state ($\boldsymbol{z}$).
\item Linear layer: This layer utilizes fully connected neurons with linear activation functions to emulate the linear/bilinear Koopman dynamics in the KEF space. The weights of these neurons are $\boldsymbol{A}$ and $\boldsymbol{B}$ matrices that represent the linear/bilinear Koopman dynamics.
\item Projection network: This network consists of fully connected linear neurons that project the high-dimensional representation back to the original state space. The weights of this layer represent the matrix $\boldsymbol{C}$.
\end{itemize}

 It is important to note that the architecture of the linear Koopman model is very similar to that of the bilinear one. The key difference lies in the linear layer, which is designed to capture linear dynamics instead of bilinear dynamics of lifted states. Hence, for the linear Koopman network, the linear layer consists of $m$ input neurons with $u_i$ as input to $i^{th}$ input neuron. In contrast, for the bilinear Koopman network, the linear layer consists of $pm$ input neurons with $z_i u_j$ as inputs for $i=1,..,p.\; j = 1,...,m$. This fundamental difference in the linear layer design persists across both the nominal Koopman network and the online adaptive network. \par

The offline training process for the nominal Koopman architecture leverages a composite loss function that incorporates the contributions of three distinct loss terms- a) reconstruction loss, b) prediction loss, and c) lifted-state prediction loss. These terms are designed based on the functionality of different components of the network. A detailed explanation of these losses can be found in \cite{sah2024real}. Mathematically, these losses are defined as:
\begin{eqnarray}
    \label{eq:offline_nn_loss}
    L_{rec} &=& \| \boldsymbol{x}_k {-} \boldsymbol{C} \boldsymbol{z}_k \|_2, \;L_{pred} = \| \boldsymbol{x}_{k+1} {-} \boldsymbol{C} \boldsymbol{\hat{z}}_{k+1} \|_2, \nonumber \\
    L_{lift} &=& \| \boldsymbol{z}_{k+1} {-} \boldsymbol{\hat{z}}_{k+1} \|_2,
\end{eqnarray}

The total loss is the weighted sum of all the three losses:
\begin{eqnarray}
    & L_{nom} {=} \alpha_1 L_{rec} {+} \alpha_2 L_{pred} {+} \alpha_3 L_{lift} {+} \gamma_1 \|\boldsymbol{W}\|_1 {+} \gamma_2 \| \boldsymbol{W}\|_2, \nonumber
\end{eqnarray}
where, $\alpha_1$, $\alpha_2$, and $\alpha_3$ are the weighing factors. $\boldsymbol{W}$ represents the trainable weights of the system. To prevent overfitting, the total loss function additionally incorporates $L_1$ and $L_2$ regularization terms applied to the network's total trainable parameters. These regularization terms are weighted by hyperparameters $\gamma_1$ and $\gamma_2$, respectively. \par

Since the adaptive neural network learns the prediction error equations~\ref{eq:adapt_lin_equation} and \ref{eq:adapt_equation_bil}, its network structure is a replica of the linear layer of the nominal Koopman neural network, i.e., it consists of a fully connected neural network architecture with linear activation functions, as shown in Fig.~\ref{fig:adap_koop}. The adaptation network is trained by minimization of loss function defined as $L_{dif} = \| \boldsymbol{\Delta z}_{k,bil} {-} \boldsymbol{\Delta A} \boldsymbol{z}_{k-1} {-} \boldsymbol{ \Delta B}(\boldsymbol{z}_{k-1}\otimes\boldsymbol{u}_{k-1})\|_2$ and $L_{dif} = \| \boldsymbol{\Delta z}_{k,lin} {-} \boldsymbol{\Delta A} \boldsymbol{z}_{k-1} {-} \boldsymbol{ \Delta B}{u}_{k-1})\|_2$ for the bilinear and linear architectures, respectively. The overall composite loss is defined as:
\begin{eqnarray}
\label{eq:adapt_loss}
L_{adapt} = L_{dif} +\beta_1\|\boldsymbol{\Delta A}\|_1   + \beta_2\| \boldsymbol{\Delta B}\|_1 \nonumber \\ +\beta_3 \| \boldsymbol{\Delta A}\|_2 + \beta_4 \| \boldsymbol{\Delta B}\|_2.
\end{eqnarray}

Here, $L_{dif}$ quantifies the discrepancy between the predicted lifted state $\boldsymbol{\hat{z}}_{k}$ and the observed lifted state $\boldsymbol{z}_{k,obs}$. The remaining terms enforce regularization on the trainable weights $\boldsymbol{\Delta A} \text{ and } \boldsymbol{\Delta B}$ of the adaptive network using $L_1 (\beta_1, \beta_2)=\beta_1\|\boldsymbol{\Delta A}\|_1   + \beta_2\| \boldsymbol{\Delta B}\|_1$ and $L_2 (\beta_3, \beta_4)=\beta_3 \| \boldsymbol{\Delta A}\|_2 + \beta_4 \| \boldsymbol{\Delta B}\|_2$ penalties. As shown in Table~\ref{tab:nn_params}, the regularization factors $\beta_1, \beta_2, \beta_3, \text{and } \beta_4$ for training the adaptive network are typically larger compared to those used for the offline model. This suggests that the elements within the matrices $\boldsymbol{\Delta A}$ and $\boldsymbol{\Delta B}$ are sparse. This implies that even though the disturbances/uncertainties are nonlinear, they can be effectively described by a limited number of interactions between the lifted state variables in the Koopman invariant subspace. Consequently, the adaptive Koopman architecture can represent these nonlinearities within a linear or bilinear framework, thereby preserving the existing control policy design. This offers a significant advantage over traditional methods that estimate model uncertainties in the original state space.  Such approaches introduce the learned uncertainties as nonlinear terms or disturbances, necessitating the development of dedicated nonlinear controllers. \par

The hyperparameters of the neural networks employed in all three study examples are detailed in Table~\ref{tab:nn_params}. All network architectures are implemented using the PyTorch library. In this work, the activation function is linear for linear neurons and $tanh(\cdot)$ for all others.

\begin{table}[ht!]
\caption{\label{tab:nn_params} Hyperparameters of Koopman network}
    \begin{tabular}{ |p{2 cm}|p{1.75 cm}|p{1.75cm}|p{1.75 cm}| } 
     \hline
      & {Coupled Pendulum} & {3R manipulator} & {Planar Quadrotor} \\ 
     \hline
     \multicolumn{4}{|c|} {Nominal Koopman Autoencoder (linear and bilinear)}\\
     \hline
    {Architecture} & [10, 40, 40, 17] & [6, 30, 30, 10] & [4, 20, 20, 10] \\ \hline
    { \# lifted state} & 17 & 17 & 15 \\\hline
     {$\alpha_1$}, {$\alpha_2$}, {$\alpha_3$} & 1, 0.4, 1 & 1, 0.3, 1 & 1, 0.3, 1\\  \hline
     {$\gamma_1$}, {$\gamma_2$} & $1e-5$, $1e-5$ & 0, 0 & 0, 0\\ \hline
     {Batch Size} & 256 & 256 & 256\\
     \hline
      \multicolumn{4}{|c|} {Koopman Adaptive Architecture}\\
      \hline
     & linear  & linear & linear \\ 
     \hline
    { \# lifted state} & 17 & 17 & 15 \\\hline
    {activation function} & linear &linear &linear \\\hline
     {$\beta_1(=\beta_2)$}, {$\beta_3(=\beta_4)$} & $0.05, 0.05 $ & 1, 0.01 & 1, 1\\ \hline
    {window size} & 4 & 10 & 10\\ \hline
     {Epochs} & 2 & 10 & 10\\ \hline
     {Optimizer} & adam & adam & adam \\  
       
     \hline
    \end{tabular}

\end{table}

In Table \ref{tab:nn_params}, the architecture of the coupled pendulum is written as $[10,40,40,17]$, which basically means that the encoder has an input layer of dimension $10$ corresponding to each state, there are two hidden layers, each of width $40$ and the output dimension is $17$, which basically is the dimension of the lifted states. The same nomenclature holds for the other two systems. Note that it is general practice to include the original states $\boldsymbol{x}$ as a part of the lifted states for ease of control implementation. This is reflected in the projection matrix $\boldsymbol{C}$, which is an identity matrix of dimension $n$ augmented with a zero matrix of dimension $n \times (p-n)$, where $p$ is the dimension of the lifted states.

\subsection{Generating Dataset for Offline Training}
To train and validate the nominal Koopman model, datasets are generated by numerically solving the system's governing equations using Python libraries. The dataset used for learning the nominal model consists of a collection of $M$ trajectory of length $T$. Each trajectory consists of $N$ data points with the state and control input sampled at fixed sampling intervals $\Delta t$. The dataset obtained as a result is given as
\begin{eqnarray}
    &\boldsymbol{\mathcal{D}}_{nom} = \left(\boldsymbol{X}_i, \boldsymbol{Y}_i, \boldsymbol{U}_i \right)_{i=1}^M,\,\boldsymbol{U}_i = \begin{bmatrix}
      \boldsymbol{u}_{1}^{i}, \boldsymbol{u}_{2}^{i},\dots, \boldsymbol{u}_{N-1}^{i}  
    \end{bmatrix},\nonumber\\
    &\boldsymbol{X}_i = \begin{bmatrix}
      \boldsymbol{x}_{1}^{i}, \boldsymbol{x}_{2}^{i}, \dots, \boldsymbol{x}_{N-1}^{i}  
    \end{bmatrix},\, 
    \boldsymbol{Y}_i = \begin{bmatrix}
      \boldsymbol{x}_{2}^{i}, \boldsymbol{x}_{3}^{i}, \dots, \boldsymbol{x}_{N}^{i}  
    \end{bmatrix}.\nonumber   
\end{eqnarray}

 For the coupled pendulum system, random initial states are chosen within the workspace given by $ \theta \in [-20,20]$ rad and $\dot{\theta} \in [-1,1]$ rad/s. For each initial state, a unique trajectory is generated by applying random, continuous input signals, allowing the generation of rich and diverse datasets. For the serial manipulator and the planar quadrotor, initially, random points are sampled within the system's output space. Then, a smooth trajectory is obtained with the sampled points used as waypoints via polynomial curves. Then, any nominal controller can be used to follow the trajectories generated to obtain the state and control snapshots. We generate 100 trajectories for each system. The number of snapshots is chosen to be $800$, $100$, and $600$ for the coupled pendulum, the 3R manipulator, and the planar quadrotor, respectively. The time-step between each snapshot is $0.01$ s for all the systems. The generated dataset is then split, with $75 \%$ allocated for training the model and the remaining $25 \%$ reserved for validation purposes. 


\subsection{Training for Online Adaptation}\label{sec:online_adap}
 The online adaptation framework employs a continuous learning paradigm where the online adaptive network is trained on a sequence of real-time datasets. Intuitively, the weights of the online adaptation network $\boldsymbol{\Delta A}$, $\boldsymbol{\Delta B}$ are initialized to zero. The adaptation module can be trained using two primary paradigms: single-step and multi-step adaptation. In single-step adaptation, the error between the predicted lifted state, $\boldsymbol{\hat{z}}_{k}$, and the lifted state corresponding to the observed system state, denoted by $\boldsymbol{x}_{k,obs}$ at time step $k$, is utilized to update the network parameters. However, this approach can lead to significant weight fluctuations due to its reliance on error information from a single time step. This work leverages multi-step adaptation, which incorporates error information from a window of $w$ latest time steps for network training. At time step $k$, the error between a batch of $w$ past observed states and their corresponding predicted states are used for training. The complete dataset employed for online learning at $k^{th}$ time step is given as
\begin{eqnarray}
    &\boldsymbol{\mathcal{D}}_{k, online} = \left(\boldsymbol{Z}, \boldsymbol{\Delta Z}, \boldsymbol{U} \right), \, \boldsymbol{\Delta {Z}}_{k} = \left[\boldsymbol{\Delta {z}}_{k-w+1}, \hdots , \boldsymbol{\Delta {z}}_{k}\right],\nonumber \\
    &\boldsymbol{Z}_{k,obs} = \left[\boldsymbol{z}_{k-w,obs}, \hdots, \boldsymbol{z}_{k-1,obs} \right],\,\boldsymbol{U}_{k} = \left[\boldsymbol{u}_{k-w}, \hdots , \boldsymbol{u}_{k-1} \right]. \nonumber  
\end{eqnarray}
where, $\boldsymbol{\Delta z}_{i} = \boldsymbol{z}_{i,obs} - \boldsymbol{\hat{z}}_{i}$, where $\boldsymbol{z}_{i,obs}$ and $\boldsymbol{\hat{z}}_{i}$ are the observed and predicted value of lifted state for $i^{th}$ time-step. This approach provides a richer dataset encompassing recent predicted and actual states compared to single-step training. By incorporating this batch of data, multi-step adaptation facilitates more robust and stable training of the adaptation network, ultimately leading to improved performance.


\subsection{Model Predictive Control (MPC)}
\label{sec:MPC}
In this study, the proposed adaptive Koopman architecture (\ref{eq:adap_lin}) is paired with the linear MPC, the implementation of which is formulated as the solution to the optimization problem with a prediction (and control) horizon $S$ as 

\begin{align}
    \label{eq:linear_MPC}
    &\min_{Z,U} \sum_{k=0}^{S}\begin{bmatrix}
     \boldsymbol{C}\boldsymbol{z}_k - \boldsymbol{x}_{ref,k}\\
     \boldsymbol{u}_k
    \end{bmatrix}^T \boldsymbol{R} \begin{bmatrix}
     \boldsymbol{C}\boldsymbol{z}_k - \boldsymbol{x}_{ref,k}\\
     \boldsymbol{u}_k
    \end{bmatrix} \\
    &\text{s.t} \,\, \boldsymbol{z}_{k{+}1} {=} \boldsymbol{A}\boldsymbol{z}_k {+} \boldsymbol{B}\boldsymbol{u}_k, \,\boldsymbol{z}_0 = \boldsymbol{\varphi}(\boldsymbol{x}_0),\nonumber\\
    &\boldsymbol{x}^{-} \leq \boldsymbol{C}\boldsymbol{z}_k \leq \boldsymbol{x}^{+}, \, \boldsymbol{u}^{-} \leq \boldsymbol{u}_k \leq \boldsymbol{u}^{+} ,\, k = 0,\, ...,\, S-1, \nonumber 
\end{align}
where $\boldsymbol{x}_{ref,k}$ refers to the reference trajectory at $k^{th}$ timestep. $\boldsymbol{R}$ represents the penalty matrix, while $[\boldsymbol{x}^{-},\,\boldsymbol{x}^{+}]$ and $[\boldsymbol{u}^{-},\,\boldsymbol{u}^{+}]$ represent the constraints on the state and control input, respectively.

Similarly, the bilinear MPC optimization for the bilinear model~(\ref{eq:adap_bil}) is given as

\begin{align}
    \label{eq:bilinear_MPC}
    &\min_{Z,U} \sum_{k=0}^{S}\begin{bmatrix}
     \boldsymbol{C}\boldsymbol{z}_k - \boldsymbol{x}_{ref,k}\\
     \boldsymbol{u}_k
    \end{bmatrix}^T \boldsymbol{R} \begin{bmatrix}
     \boldsymbol{C}\boldsymbol{z}_k - \boldsymbol{x}_{ref,k}\\
     \boldsymbol{u}_k
    \end{bmatrix} \\
    &s.t \, \boldsymbol{z}_{k{+}1} {=} \boldsymbol{A}\boldsymbol{z}_k {+} \boldsymbol{B}(\boldsymbol{z}_k\otimes\boldsymbol{u}_k),\,\boldsymbol{z}_0 = \boldsymbol{\varphi}(\boldsymbol{x}_0), \nonumber\\
    &\boldsymbol{x}^{-} \leq \boldsymbol{C}\boldsymbol{z}_k \leq \boldsymbol{x}^{+}, \, \boldsymbol{u}^{-} \leq \boldsymbol{u}_k \leq \boldsymbol{u}^{+} ,\, k = 0,\, ...,\, S-1. \nonumber
\end{align}

The linear and bilinear MPC schemes are implemented via Sequential Quadratic Programming (SQP) that relies on the OSQP (Operator Splitting QP) solver~\cite{osqp}, in a manner similar to~\cite{folkestad2022koopnet}. 

\subsection{Data Availability}
The data used in the study can be found at GitHub: \href{https://github.com/Rajpal9/Adaptive-koopman/tree/main/datasets}{https://github.com/Rajpal9/Adaptive-koopman/tree/main/datasets}
\subsection{Code Availability}
The code used for generating all the results in the study can be found at GitHub:  \href{https://github.com/Rajpal9/Adaptive-koopman}{https://github.com/Rajpal9/Adaptive-koopman}



\section{Authorship Contribution}
All three authors contributed to the conception, design and development of the proposed algorithm. R.S and C.K.S. implemented, tested, and refined
the method. The manuscript was written by R.S and C.K.S with support from J.K.
\section{Competing Interests}
The authors have no competing interests to declare.



\end{document}